\documentclass[letterpaper]{article} 
\usepackage[]{aaai24}  
\usepackage{times}  
\usepackage{helvet}  
\usepackage{courier}  
\usepackage[hyphens]{url}  
\usepackage{graphicx} 
\usepackage{natbib}  
\usepackage{caption} 
\newcommand{\longversion}[1]{}
\newcommand{\shortversion}[1]{#1}
\newcommand{\futuresketch}[1]{}
\usepackage{times}
\usepackage{soul}
\usepackage{url}
\usepackage[utf8]{inputenc}
\usepackage{multirow}
\usepackage{amssymb,amsfonts,amsmath,amsthm}%

\usepackage{enumitem}
\usepackage{stackengine}
\usepackage{calc}
\newlength\shlength
\newcommand\xshlongvec[2][0]{\setlength\shlength{#1pt}%
  \stackengine{-5.6pt}{$#2$}{\smash{$\kern\shlength%
    \stackengine{7.55pt}{$\mathchar"017E$}%
      {\rule{\widthof{$#2$}}{.57pt}\kern.4pt}{O}{r}{F}{F}{L}\kern-\shlength$}}%
      {O}{c}{F}{T}{S}}

\usepackage{xspace}
\usepackage[ruled,vlined,linesnumbered]{algorithm2e}
\usepackage[table,x11names]{xcolor}
\usepackage{graphicx}  
\usepackage{caption}

\SetKwInput{KwData}{In}
\SetKwInput{KwResult}{Out}
\SetAlFnt{\small}
\SetAlCapFnt{\small}
\SetAlCapNameFnt{\small}
\SetAlCapHSkip{0pt}
\SetEndCharOfAlgoLine{}

\makeatother

\makeatletter
\newcommand*{\inlineequation}[2][]{%
  \begingroup
    \refstepcounter{equation}%
    \ifx\\#1\\%
    \else
      \label{#1}%
    \fi
    \relpenalty=10000 %
    \binoppenalty=10000 %
    \ensuremath{%
      #2%
    }%
    ~\@eqnnum
  \endgroup
}

\newcommand{\mtext}[1]{\ensuremath{\mathcal{#1}}}
\newcommand{\cntc}[0]{\ensuremath{\#\cdot}}

\DeclareMathOperator{\poly}{poly}

\DeclareMathOperator{\ord}{pos}

\newcommand{\camera}[1]{}

\newcommand{\QBFSAT}{\textsc{QSat}\xspace}

\DeclareMathOperator{\tower}{\ensuremath{\mathsf{tow}}}

\usepackage{microtype}
\usepackage{tikz}
\usetikzlibrary{shapes}
\usetikzlibrary{calc}

\newcommand{\citex}[1]{\citeauthor{#1}~\shortcite{#1}}

\renewcommand{\P}{\ensuremath{\textsc{P}}\xspace}
\newcommand{\NP}{\ensuremath{\textsc{NP}}\xspace}
\newcommand{\SIGMA}[2]{\ensuremath{\Sigma_{\textrm{#1}}^{\textrm{#2}}}}

\usetikzlibrary{positioning}

\tikzstyle{tdnode} = [draw,rounded corners,top color=vertexTopColor,bottom color=vertexBottomColor,minimum size=1.5em]
\tikzstyle{stdnode} = [tdnode, font=\scriptsize]
\tikzstyle{stdnodecompact} = [stdnode, inner sep = 1.5pt, outer sep = 0.1pt]
\tikzstyle{stdnodetable} = [stdnode, inner sep = 0.5pt, outer sep = 0]
\tikzstyle{stdnodenum} = [minimum size=1.5em, font=\scriptsize]
\tikzstyle{tdedge} = [-,draw,thick]
\tikzstyle{tdlabel} = [draw=none, rectangle, fill=none, inner sep=0pt, font=\scriptsize]
\colorlet{vertexTopColor}{white}
\colorlet{vertexBottomColor}{black!10}

\usepackage{ctable}
\usepackage{ifthen}
\newif\iflong
\usepackage{todonotes}

\usepackage{float}

\newcommand{\SB}{\{}%
\newcommand{\SM}{\mid}%
\newcommand{\SE}{\}}%

\newcommand{\ta}[1]{\ensuremath{2^{#1}}}
\newcommand{\Card}[1]{\left|#1\right|}
\newcommand{\CCard}[1]{\|#1\|}

\DeclareMathOperator{\width}{width}
\DeclareMathOperator{\children}{chldr}

\DeclareMathOperator{\rootOf}{root}

\usepackage{mathtools}
\DeclarePairedDelimiter\ceil{\lceil}{\rceil}

\newcommand{\bvali}[3]{\ensuremath{[\![#1]\!]_{#2,#3}}}
\newcommand{\bval}[2]{\ensuremath{[\![#1]\!]_{#2}}}

\makeatletter
\newcommand{\algorithmfootnote}[2][\footnotesize]{
  \let\old@algocf@finish\@algocf@finish
  \def\@algocf@finish{\old@algocf@finish
    \leavevmode\rlap{\begin{minipage}{\linewidth}
    #1#2
    \end{minipage}}
  }
}
\makeatother

\newcommand{\TTT}{\ensuremath{\mathcal{T}}}%
\newcommand{\WWW}{\ensuremath{\mathcal{W}}}%
\newcommand{\por}{\vee}

\newcommand{\eqdef}{\ensuremath{\,\mathrel{\mathop:}=}}
\newcommand{\hsep}{\leftarrow\,}
\newcommand{\SAT}{\textsc{SAT}\xspace}
\newcommand{\ASP}{\textsc{ASP}\xspace}

\newcommand{\at}{\text{\normalfont at}}
\newcommand{\var}{\text{\normalfont var}}
\newcommand{\bigO}[1]{\ensuremath{{\mathcal O}(#1)}}

\newcommand{\prog}{\ensuremath{\Pi}}

\newcommand{\Tab}[1]{\ensuremath{\text{C-Tabs}}}

\newcommand{\tw}[1]{\mathit{tw}(#1)}

\newcommand{\Nat}{\mathbb{N}} %
\DeclareMathOperator{\type}{type}

\newcommand{\leaf}{\textit{leaf}}
\newcommand{\inner}{\textit{inner}}

\newcommand{\join}{\textit{join}}

\newtheorem{example}{Example}

\newtheorem{proposition}{Proposition}

\newtheorem{theorem}{Theorem}
\newtheorem{lemma}{Lemma}
\newtheorem{definition}{Definition}
\newtheorem{corollary}{Corollary}

{%
  \addtocounter{observation}{-1}
  \endgroup
}%

\newenvironment{restatecorollary}[1][\unskip]{%
  \begingroup
  \def\thecorollary{\ref{#1}} 
}%
{%
  \addtocounter{corollary}{-1}
  \endgroup
}%

{%
  \addtocounter{proposition}{-1}
  \endgroup
}%

\newenvironment{restatetheorem}[1][\unskip]{%
  \begingroup
  \def\thetheorem{\ref{#1}} 
}%
{%
  \addtocounter{theorem}{-1}
  \endgroup
}%

\newenvironment{restatelemma}[1][\unskip]{%
  \begingroup
  \def\thelemma{\ref{#1}} 
}%
{%
  \addtocounter{lemma}{-1}
  \endgroup
}%

 \title{Extended Version of: \\On the Structural Hardness of Answer Set Programming: Can Structure Efficiently Confine the Power of Disjunctions?} 

  \author{
    Markus Hecher\textsuperscript{\rm 1}, 
    Rafael Kiesel\textsuperscript{\rm 2}
}

\affiliations{
    \textsuperscript{\rm 1} Massachusetts Institute of Technology, Cambridge, USA\\
    \textsuperscript{\rm 2} TU Wien, Vienna, Austria\\
    hecher@mit.edu, 
    rkiesel@tuwien.ac.at
}

\begin{document}
%
%
\maketitle
\begin{abstract}
Answer Set Programming (\ASP) is a generic problem modeling and solving framework with a strong focus on knowledge representation and 
a rapid growth of industrial applications. 
%
%
%
So far, the study of complexity resulted in characterizing hardness and determining their sources, 
fine-grained insights
in the form of dichotomy-style results, 
as well as detailed parameterized complexity landscapes.
%
%
%
%
%
Unfortunately, for the well-known parameter treewidth disjunctive programs 
require double-exponential runtime under reasonable complexity assumptions. This quickly becomes out of reach. 
%
%
We deal with the classification of structural parameters for disjunctive \ASP on the program's rule structure (incidence graph). 

First, we provide a polynomial kernel to obtain single-exponential runtime in terms of vertex cover size, despite subset-minimization being not represented in the program's structure. Then we turn our attention to strictly better structural parameters between vertex cover size and treewidth. Here, we provide double-exponential lower bounds for the most prominent parameters in that range: treedepth, feedback vertex size, and cliquewidth. Based on this, we argue that unfortunately our options beyond vertex cover size are 
limited.
Our results provide an in-depth hardness study, relying on a novel reduction from normal to disjunctive programs, trading the increase of complexity for an exponential parameter~compression.
%
%
%
%
%
%

\end{abstract}

\section*{Introduction}

Answer Set Programming (ASP)~\cite{BrewkaEiterTruszczynski11,GebserKaminskiKaufmannSchaub12} is a prominent declarative modeling and solving framework, enabling to efficiently solve problems in knowledge representation and artificial intelligence. 
Indeed, in the last couple of years, \ASP has been extended several times and it grew into a rich modeling language, where solvers like clasp~\cite{GebserKaufmannSchaub09a} or wasp~\cite{AlvianoEtAl19} are readily available.
This makes \ASP a suitable target language for many problems, e.g., \cite{BalducciniGelfondNogueira06a,NiemelaSimonsSoininen99,NogueiraBalducciniGelfond01a,GuziolowskiEtAl13a,SchaubWoltran18,AbelsEtAl19},
where
problems are encoded in a logic program, which is a set of rules whose solutions are called answer sets.
%
%

Unfortunately, in terms of computational complexity, there is a catch to \ASP's modeling comfort.
%
%
Indeed, while there exist \NP-complete fragments~\cite{BidoitFroidevaux91,MarekTruszczynski91,Ben-EliyahuDechter94}, 
the \emph{consistency problem} of deciding whether a disjunctive program admits an answer set is inherently hard, yielding $\Sigma_2^P$-completeness~\cite{EiterGottlob95}, even
%
%
if we restrict rule body sizes to a constant~\cite{Truszczynski11}.

%
%
%
%
%
%
%
How do we deal with this high complexity?
Parameterized
complexity~\cite{CyganEtAl15,Niedermeier06,DowneyFellows13,
FlumGrohe06}, 
offers a framework, enabling to analyze a problem's hardness in terms of
certain \emph{parameter(s)}, which has been extensively applied to \ASP~\cite{GottlobScarcelloSideri02,GottlobPichlerWei10,LacknerPfandler12,FichteKroneggerWoltran19}. 
%
%
For \ASP there is growing research on the prominent structural parameter \emph{treewidth}~\cite{JaklPichlerWoltran09,CalimeriEtAl16,BichlerMorakWoltran18,BliemEtAl20,EiterHecherKiesel21}. 
Intuitively, the measure treewidth enables the solving of numerous combinatorially hard problems in parts. This parameter indicates the maximum number of variables of these parts one has to investigate during problem solving.
%
Unfortunately, there is still a catch: Under the \emph{Exponential Time Hypothesis (ETH)}~\cite{ImpagliazzoPaturiZane01}, the evaluation of disjunctive programs is prohibitively expensive, being inherently \emph{double exponential} in its treewidth~\cite{Hecher22}. 
%
Indeed this is a bad worst case; $2^{2^k}$ is larger than the (est.) number of atoms in the universe, already for $k{=}9$.

This motivates our search for preferably small structural parameters that enable single-exponential runtimes, i.e., we aim for structural properties strong enough to be significantly more exploitable than treewidth.
We thereby consider the program's rule structure (incidence graph representation) 
and we only study parameters smaller than the vertex cover size, which
is the number of vertices needed to cover every edge of a graph.
We focus on the following structural parameters that are \emph{more generally} applicable than treewidth, 
but smaller than vertex cover size, and state their intuitive meaning\footnote{For formal definitions, see the preliminaries.}.
%
\begin{itemize}[leftmargin=5pt, itemindent=8pt, itemsep=-1pt]
	\item Treedepth: How close is the structure to being a star?
	\item Feedback vertex number: How many atoms or rules need to be removed to obtain an acyclic graph structure?
	\item Pathwidth: How close is the structure to being a path?
\end{itemize}
Our results will also have consequences for the parameters
%
\begin{itemize}[leftmargin=5pt, itemindent=8pt, itemsep=-1pt]
	\item Bandwidth: How ``far'' do structural dependencies reach, assuming atoms and rules are linearly ordered?
	\item Cutwidth: How well can we linearly order atoms and rules, aiming for minimizing structural dependencies between predecessors of an atom/rule and their successors?
	\item Cliquewidth: How close is the structure to being a clique?
\end{itemize}
%
%
%
This paper thereby asks the following:
%
%
%
%
\begin{itemize}[leftmargin=5pt, itemindent=8pt, itemsep=-1pt]
	\item Can we evaluate disjunctive \ASP in single-exponential time for structural measures more general than treewidth? 
	%
	%
%
	\item 
	What makes a disjunctive program's structure harder to exploit than the structure of normal programs? 
	\item Can we leverage the hardness of disjunction by translating from normal programs, thereby trading for an exponential decrease of structural dependencies (parameters)?
\end{itemize}

\smallskip\noindent\textbf{Contributions.} We address these questions via algorithms and novel reductions, thereby establishing the following.
\begin{itemize}[itemsep=-1pt]
	\item First, we show a single-exponential upper bound when considering vertex cover size as a parameter. 
	This is challenging, since our algorithm works despite the implicit subset-minimization required by disjunctive \ASP, which is, however not directly manifested in the structural representation of the program.
	We present a polynomial-sized kernel for disjunctive \ASP, which then
	yields a single-exponential algorithm for computing answer sets.
	%
	%
	%
	%
%
	\item Then, we show how to reduce from  normal \ASP to (full) disjunctive \ASP,
	thereby exponentially \emph{decreasing the program's cyclicity} (feedback vertex size)  from~$k$ to~$\log(k)$.
%
%
%
To the best of our knowledge, this is the first reduction that exponentially reduces structural dependency of logic programs,
at the cost of solving a harder program fragment.
This reduction leads to ETH-tight double-exponential lower bounds for feedback vertex size.\footnote{The results even hold for 
the smallest number of atoms that, upon removal from the program, yields a structure of 
 (almost)~paths.}
\item 
%
 %
 The idea of this reduction technique has many further consequences. 
 Indeed, by a generalization of this concept we obtain tight double-exponential bounds for the structural parameter treedepth.
 Even further, with this idea, we rule out single-exponential algorithms for the structural measures
pathwidth, bandwidth, cliquewidth, and cutwidth.
 Unfortunately, given these lower bounds for most prominent measures between vertex cover and treewidth as well as the observations of our algorithm for vertex cover 
	we do not expect significantly better parameters yielding better runtimes -- disjunction limits structural exploitability.
%
%
%
%
	
\end{itemize}

\smallskip\noindent\textbf{Related Work.} 
\futuresketch{
For normal and HCF programs, slightly superexponential algorithms in the treewidth~\cite{FichteHecher19} for solving consistency are known,
which has been improved for
%
so-called~$\iota$-tight programs~\cite{FandinnoHecher21}.} 
%
%
For disjunctive ASP,
algorithms have been proposed~\cite{JaklPichlerWoltran09,PichlerEtAl14} 
running in time linear in the instance size, but double exponential
in the treewidth. 
Hardness of problems has been studied by means of runtime dependency in the treewidth, e.g., levels of 
exponentiality, where triple-exponential algorithms are known~\cite{LokshtanovMarxSaurabh11,
MarxMitsou16,FichteHecherPfandler20}.
For quantified Boolean formulas (QBFs) parameterized by vertex cover size, a single-exponential runtime result is known~\cite{LampisMitsou17}. However, this result does not trivially transfer to \ASP, as a direct encoding of subset-minimization causes an unbounded increase of vertex cover size.
Also, lower bounds for QBFs and some of our considered parameters are known~\cite{PanVardi06,LampisMitsou17,FichteHecherPfandler20,FichteEtAl23}. We strengthen these results for \ASP, providing ETH-tight bounds for many measures.
Programs of bounded even or odd cycles have been analyzed~\cite{LinZhao04}. 
%
%
%
%
Further, the 
feedback width has been studied, which depends on the atoms required to break large SCCs~\cite{GottlobScarcelloSideri02}.
%
%
%
%
%
\futuresketch{The proposed algorithm 
was used for 
counting answer sets involving projection~\cite{GebserKaufmannSchaub09a}, 
which is at least double exponential~\cite{FichteEtAl18} in the treewidth. 
However, for plain counting (single exponential), it can overcount due to lacking unique level mappings (orderings). 
}
For \SAT, empirical results~\cite{AtseriasFichteThurley11}
involving resolution-width 
and treewidth yield efficient \SAT solver runs on instances of small treewidth.

\section*{Preliminaries}
We assume familiarity with propositional satisfiability (\SAT)~\cite{BiereHeuleMaarenWalsh09,KleineBuningLettman99}. 
%

\futuresketch{
\paragraph{Basics and Combinatorics.}
For a set~$X$, let $\ta{X}$ be the \emph{power set of~$X$}
consisting of all subsets~$Y$ with $\emptyset \subseteq Y \subseteq X$.
Let $\vec s$ be a sequence of elements of~$X$. When we address the
$i$-th element of the sequence~$\vec s$ for a given positive
integer~$i$, we simply write $\vec s_{(i)}$. The sequence~$\vec s$
\emph{induces} an \emph{ordering~$<_{\vec s}$} on the elements in~$X$
by defining the
relation~$<_{\vec s} \eqdef \SB (\vec s_{(i)},\vec s_{(j)}) \SM 1 \leq
i < j \leq \Card{\vec s}\SE$.
Given some integer~$n$ and a family of finite subsets~$X_1$, $X_2$,
$\ldots$, $X_n$. Then, the generalized combinatorial
inclusion-exclusion principle~\cite{GrahamGrotschelLovasz95a} states
that the number of elements in the union over all subsets is
$\Card{\bigcup^n_{j = 1} X_j} = \sum_{I \subseteq \{1, \ldots, n\}, I
  \neq \emptyset} (-1)^{\Card{I}-1} \Card{\bigcap_{i \in I} X_i}$.
}

\smallskip\noindent\textbf{Answer Set Programming (ASP).}
%
%
%
 %
We follow standard definitions of propositional ASP~\cite{BrewkaEiterTruszczynski11,JanhunenNiemela16a}.
%
Let $\ell$, $m$, $n$ be non-negative integers such that
$\ell \leq m \leq n$, $a_1$, $\ldots$, $a_n$ be distinct propositional
atoms. Moreover, we refer by \emph{literal} to an atom or the negation
thereof.
%
A \emph{program}~$\prog$ is a set of \emph{rules} of the form
%
\(
a_1\por \cdots \por a_\ell \hsep a_{\ell+1}, \ldots, a_{m}, \neg
a_{m+1}, \ldots, \neg a_n.
\)
%
%
%
%
%
%
%
%
%
%
For a rule~$r$, we let $H_r {\eqdef} \{a_1, \ldots, a_\ell\}$,
$B^+_r {\eqdef} \{a_{\ell+1}, \ldots, a_{m}\}$, and
$B^-_r {\eqdef} \{a_{m+1}, \ldots, a_n\}$.
%
%
%
We denote the sets of \emph{atoms} occurring in a rule~$r$ or in a
program~$\prog$ by $\at(r) {\eqdef} H_r \cup B^+_r \cup B^-_r$ and
$\at(\prog){\eqdef} \bigcup_{r\in\prog} \at(r)$.
Further, $\ord(r,x)$ refers to the unique \emph{position} of~$x\in\at(r)$ in~$r$.
%
%
The rules, whose head contain $a\in\at(\Pi)$ are given by~$\mathcal{H}(a){\eqdef}\{r{\,\in\,} \Pi \mid a{\,\in\,} H_r\}$.
We define the completion~$\mathcal{C}(\Pi)$  by~$\Pi \cup \{r_1^h \vee \ldots \vee\allowbreak r_\ell^h \leftarrow h \mid h{\,\in\,} \at(\Pi), \mathcal{H}(h){\,=\,}\{r_1, \ldots, r_\ell\}\} \cup \{\leftarrow r^h, \neg a,\,\, \leftarrow r^h, b,\allowbreak 
\mid r\in \Pi, 
a\in B_r^+, b\in B_r^- \cup H_r, b\neq h\}$~\cite{Clark77}.
%
%
Program~$\prog$ is \emph{normal} if $\Card{H_r} \leq 1$ for
every~$r \in \prog$; $\prog$ is \emph{normalized}
if for every~$r\in\prog$, $\Card{H_r\cup B_r^+\cup B_r^-}\leq 3$. Normalization preserves hardness for known fragments~\cite{Truszczynski11}.
The \emph{dependency graph}~$D_\prog$ of $\prog$ is the
directed graph on the atoms~$\bigcup_{r\in \prog}H_r \cup B^+_r$, where for every
rule~$r \in \prog$, two atoms $a\in B^+_r$ and~$b\in H_r$ are joined by
an edge~$(a,b)$.
Program~$\Pi$ is \emph{tight} if~$D_\prog$ is acyclic.
\futuresketch{A head-cycle of~$D_\prog$ is an $\{a, b\}$-cycle\footnote{Let
  $G=(V,E)$ be a digraph and $W \subseteq V$. Then, a cycle in~$G$ is
  a $W$-cycle if it contains all vertices from~$W$.} for two distinct
atoms~$a$, $b \in H_r$ for some rule $r \in \prog$. 
Program~$\prog$ is
\emph{head-cycle-free} if $D_\prog$ contains no
head-cycle~\cite{Ben-EliyahuDechter94}.}
%
%

An \emph{interpretation} $I$ is a set of atoms. $I$ \emph{satisfies} a
rule~$r$ if $(H_r\,\cup\, B^-_r) \,\cap\, I \neq \emptyset$ or
$B^+_r \setminus I \neq \emptyset$.  $I$ is a \emph{model} of $\prog$
if it satisfies all rules of~$\prog$, in symbols $I \models \prog$. 
%
%
The \emph{Gelfond-Lifschitz
  (GL) reduct} of~$\prog$ under~$I$ is the program~$\prog^I$ obtained
from $\prog$ by first removing all rules~$r$ with
$B^-_r\cap I\neq \emptyset$ and then removing all~$\neg z$ where
$z \in B^-_r$ from the remaining
rules~$r$~\cite{GelfondLifschitz91}. %
$I$ is an \emph{answer set} of a program~$\prog$, denoted~$I\models \prog$, if $I$ is a minimal
model of~$\prog^I$. %
%
%
%
A tight program $\Pi$ is \emph{fully tight} if for every model~$M$ of~$\Pi$,
 there is a unique model~$M'$ of~$\mathcal{C}(\Pi)$ and vice versa, such that $M{\,=\,}M'\cap \at(\Pi)$.
Intuitively, this means that for computing answer sets it is sufficient to compute the models of~$\Pi$.
%
%
Deciding whether an \ASP program has an answer set is called
\emph{consistency}, which is \SIGMA{2}{P}-complete~\cite{EiterGottlob95}. 
If the input is restricted to normal programs, the complexity drops to
\NP-complete~
\cite{MarekTruszczynski91}.
\futuresketch{A head-cycle-free program~$\prog$ 
can be translated into a normal program in polynomial
time~\cite{Ben-EliyahuDechter94}.}
\futuresketch{This characterization vacuously extends to head-cycle-free
programs by results of~\citex{Ben-EliyahuDechter94}.}
%
%
%
%
%
\futuresketch{Given a program~$\prog$. It is folklore that an atom~$a$ of any answer
set of $\prog$ has to occur in some head of a rule
of~$\prog$~\cite[Ch~2]{GebserKaminskiKaufmannSchaub12}, which we 
hence assume in the following. 
}
%
%
%
%
%
%
%
\longversion{%
} 
%

%
%
%
%
%
%
%
%
%
%
%
%
%
%
%
%
%
%
\longversion{%
}%
\longversion{%
}%

\smallskip
\noindent\textbf{Graph Representations.} 
%
We assume familiarity with standard notions in computational
complexity~\cite{Papadimitriou94} and  graph theory, cf.,~\cite{Diestel12}.
%
%
%
%
For parameterized complexity, we refer to standard
texts~\cite{CyganEtAl15}. 

We need graph representations to deploy structural measures~\cite{JaklPichlerWoltran09}.
The \emph{primal graph 
} $\mathcal{G}_\prog$
of a  program~$\prog$ has the atoms of~$\prog$ as vertices and an
edge~$\{a,b\}$ if there exists a rule~$r \in \prog$ with $a,b \in \at(r)$.
%
%
%
The \emph{incidence graph 
}$\mathcal{I}_\prog$
of~$\prog$ has as vertices the atoms and rules of~$\prog$ and an
edge~$\{a,r\}$ for every rule~$r \in \prog$ with $a \in \at(r)$.

\begin{figure}[t]
\centering
	\begin{tikzpicture}[node distance=7mm,every node/.style={circle,inner sep=2pt}]
\node (b) [fill,label={[text height=1.5ex]left:$a$}] {};
\node (a) [fill,right=of b,label={[text height=1.5ex,yshift=0.0cm,xshift=-0.05cm]right:$d$}] {};
\node (c) [fill,below of=b,label={[text height=1.5ex,yshift=0.09cm,xshift=0.05cm]left:$b$}] {};
\node (d) [fill,below of=a,label={[text height=1.5ex,yshift=0.09cm,xshift=-0.05cm]right:$c$}] {};
\draw (a) to (b);
\draw (a) to (d);
\draw (b) to (c);
\draw (b) to (d);
\draw (c) to (d);
%
%
%
%
%
\node (b) [right=5.5em of d,yshift=.5em,fill,label={[text height=1.5ex,yshift=0.5em]below:$r_1$}] {};
\node (a) [fill,right=1.5em of b,label={[text height=1.5ex,yshift=0.5em,xshift=-0.00cm]below:$r_2$}] {};
\node (d2) [fill,right=1.5em of a,label={[text height=1.5ex,yshift=0.5em,xshift=-0.00cm]below:$r_3$}] {};
\node (c) [fill,right=1.5em of d2,label={[text height=1.5ex,yshift=0.5em,xshift=-0.00cm]below:$r_4$}] {};
\node (vb) [draw=black,fill=white,above=.6em of b,xshift=-0em,yshift=.0em,fill,label={[text height=1.5ex,yshift=0.35em,xshift=.3em]left:$b$}] {};
\node (va) [draw=black,fill=white,right=1.5em of vb,yshift=.0em,fill,label={[text height=1.5ex,yshift=0.35em,xshift=.3em]left:$a$}] {};
\node (vc) [draw=black,fill=white,right=1.5em of va,yshift=.0em,fill,label={[text height=1.5ex,yshift=0.35em,xshift=.3em]left:$c$}] {};
\node (vd) [draw=black,fill=white,right=1.5em of vc,yshift=.0em,fill,label={[text height=1.5ex,yshift=0.35em,xshift=.3em]left:$d$}] {};
\draw (b) to (vb);
\draw (b) to (va);
%
\draw (a) to (vb);
\draw (a) to (va);
\draw (a) to (vc);
\draw (d2) to (vd);
\draw (d2) to (va);
%
\draw (c) to (vd);
\draw (c) to (va);
\draw (c) to (vc);
%
\end{tikzpicture}
\caption{(Left): The primal graph~$\mathcal{G}_{\Pi_1}$ of program~$\Pi_1$ of Example~\ref{ex:running1}. (Right):  The incidence graph~$\mathcal{I}_{\Pi_1}$ of~$\Pi_1$.
}\label{fig:graphs}\label{fig:primal}
\end{figure}
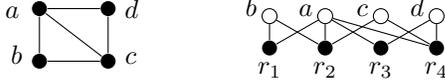

\begin{example}\label{ex:run}\label{ex:running1}
Consider the tight program~$\Pi_1 {=} \{r_1, r_2, r_3, r_4\}$,
where $r_1= b{\,\leftarrow\,} \neg a$; $r_2=b\leftarrow a, c$; $r_3=a \vee d\leftarrow$; and
$r_4=c\leftarrow a, \neg d$.
Then, $\Pi_1$ admits two answer sets~$\{b,d\}, \{a,b,c\}$.
Figure~\ref{fig:graphs} shows graph representations of~$\Pi_1$.
Program~$\Pi_2=\Pi_1 \cup \{r_1^b \vee r_2^b  \leftarrow b; \leftarrow r_1^b, a; \leftarrow r_2^b, \neg a; \leftarrow r_2^b, \neg c; \leftarrow a, d; \leftarrow c, \neg a; \leftarrow c, d\}$
is \emph{fully tight}. 
\end{example}

\futuresketch{
%
We recall some basic notions.
Let $\Sigma$ and $\Sigma'$ be some finite alphabets.  We call
$I \in \Sigma^*$ an \emph{instance} and $\CCard{I}$ denotes the size
of~$I$.  
%
Let $L \subseteq \Sigma^* \times \Nat$ and
$L' \subseteq {\Sigma'}^*\times \Nat$ be two parameterized problems. An
\emph{fpt-reduction} $r$ from $L$ to $L'$ is a many-to-one reduction
from $\Sigma^*\times \Nat$ to ${\Sigma'}^*\times \Nat$ such that for all
$I \in \Sigma^*$ we have $(I,k) \in L$ if and only if
$r(I,k)=(I',k')\in L'$ such that $k' \leq g(k)$ for a fixed computable
function $g: \Nat \rightarrow \Nat$, and there is a computable function
$f$ and a constant $c$ such that $r$ is computable in time
$O(f(k)\CCard{I}^c)$. If additionally~$g$ is a linear function,
then~$r$ is referred to as~\emph{fpl-reduction}.
%
%
%
A \emph{witness function} is a
function~$\mathcal{W}\colon \Sigma^* \rightarrow 2^{{\Sigma'}^*}$ that
maps an instance~$I \in \Sigma^*$ to a finite subset
of~${\Sigma'}^*$. We call the set~$\WWW(I)$ the \emph{witnesses}. A
\emph{parameterized counting
  problem}~$L: \Sigma^* \times \Nat \rightarrow \Nat_0$ is a
function that maps a given instance~$I \in \Sigma^*$ and an
integer~$k \in \Nat$ to the cardinality of its
witnesses~$\Card{\WWW(I)}$.
Let $\mtext{C}$ be a decision complexity class,~e.g., \P. Then,
$\cntc\mtext{C}$ denotes the class of all counting problems whose
witness function~$\WWW$ satisfies (i)~there is a
function~$f: \Nat_0 \rightarrow \Nat_0$ such that for every
instance~$I \in \Sigma^*$ and every $W \in \WWW(I)$ we have
$\Card{W} \leq f(\CCard{I})$ and $f$ is computable in
time~$\bigO{\CCard{I}^c}$ for some constant~$c$ and (ii)~for every
instance~$I \in \Sigma^*$ the decision problem~$\WWW(I)$ belongs to
the complexity class~$\mtext{C}$.
Then, $\cntc\P$ is the complexity class consisting of all counting
problems associated with decision problems in \NP.
Let $L$ and $L'$ be counting problems with witness functions~$\WWW$
and $\WWW'$. A \emph{parsimonious reduction} from~$L$ to $L'$ is a
polynomial-time reduction~$r: \Sigma^* \rightarrow \Sigma'^*$ such
that for all~$I \in \Sigma^*$, we
have~$\Card{\WWW(I)}=\Card{\WWW'(r(I))}$. It is easy to see that the
counting complexity classes~$\cntc\mtext{C}$ defined above are closed
under parsimonious reductions. It is clear for counting problems~$L$
and $L'$ that if $L \in \cntc\mtext{C}$ and there is a parsimonious
reduction from~$L'$ to $L$, then $L' \in \cntc\mtext{C}$.
%
%
%
}

\smallskip
\noindent\textbf{Treewidth \& Pathwidth.} %
A \emph{tree decomposition (TD)} \cite{RobertsonSeymour86} 
of a given graph~$G{=}(V,E)$ is a pair
$\TTT{=}(T,\chi)$ where $T$ is a tree rooted at~$\rootOf(T)$ and $\chi$ 
assigns to each node $t$ of~$T$ a set~$\chi(t)\subseteq V$,
called \emph{bag}, such that (i) $V=\bigcup_{t\text{ of }T}\chi(t)$, (ii)
$E\subseteq\SB \{u,v\} \SM t\text{ in } T, \{u,v\}\subseteq \chi(t)\SE$,
and (iii) for each $r, s, t\text{ of } T$, such that $s$ lies on the path
from~$r$ to $t$, we have $\chi(r) \cap \chi(t) \subseteq \chi(s)$.
For every node~$t$ of~$T$, we denote by $\children(t)$ the \emph{set of child nodes of~$t$} in~$T$.
%
We
let $\width(\TTT) {\eqdef} \max_{t\text{ of } T}\Card{\chi(t)}-1$.
The
\emph{treewidth} $\tw{G}$ of $G$ is the minimum $\width({\TTT})$ over
all TDs $\TTT$ of $G$. 
If~$T$ is a path we call~$\TTT$ a \emph{path decomposition (PD)}.
The \emph{pathwidth} is analogously defined to treewidth, but only 
over PDs~$\TTT$. 
For a node~$t \text{ of } T$, we say that $\type(t)$ is $\leaf$ if 
$t$ has
no children
; $\join$ if $t$ has exactly two children~$t'$ and $t''$ with
$t'\neq t''$; 
$\inner$ if~$t$ has a single child.
%
If for
every node $t\text{ of } T$, %
$\type(t) \in \{ \leaf, \join, \inner\}$, 
the TD is called \emph{nice}.
%
A TD can be turned into a nice TD~\cite{Kloks94a}[Lem.\ 13.1.3] \emph{without width-increase} in linear~time.

%

%

\smallskip\noindent\textbf{Bandwidth \& Cutwidth.}
Let~$G=(V,E)$ be a given graph and~$f: V \rightarrow \{1, \ldots, \Card{V}\}$
be a bijective (one-to-one) \emph{ordering} 
that uniquely assigns a vertex to an integer.
%
%
The \emph{bandwidth} of~$G$ is the minimum $\max_{\{u,v\}\in E} \Card{f(u)-f(v)}$ among every ordering $f$ for~$G$.
The \emph{cutwidth} of~$G$ is the  minimum $\max_{1\leq i\leq \Card{V}} \Card{\{\{x,y\} \in E \mid f(x) \leq i, f(y) > i\}}$ among every ordering~$f$ for~$G$.

\smallskip\noindent\textbf{Cliquewidth.}
The \emph{cliquewidth} of a graph~$G$ is the minimum number of labels needed to construct~$G$ via labeled graphs over the following $4$ operations: (1) create a new vertex with label~$\ell$, (2) disjoint union of two labeled graphs, (3) create new edges between all vertices with label~$\ell$ and those with~$\ell'$, and (4) rename label~$\ell$ to~$\ell'$.

\smallskip\noindent\textbf{Feedback Vertex Size.}
A set~$S$ is a \emph{feedback vertex set (FVS)} of a graph~$G$,
if the graph obtained from~$G$ after removing~$S$ is acyclic. 
$S$ is \emph{sparse} (for a program~$\Pi$) if (i) $G\in\{\mathcal{G}_\Pi, \mathcal{I}_\Pi\}$, and (ii) for every two atoms~$a,b\in\at(\Pi)\setminus S$, $\Pi$ contains at most one rule  over both~$a,b$.
The \emph{feedback vertex size} of~$G$ is the smallest~$\Card{S}$ among every FVS~$S$ of~$G$.

\smallskip\noindent\textbf{Treedepth.}
A \emph{Tr\'emaux tree}~$T$ of a graph~$G=(V,E)$ is a  rooted tree such that for every edge~$\{u,v\}\in E$, either~$u$ is an ancestor of~$v$ in~$T$ or vice versa.
The \emph{treedepth} of~$G$ is the minimum \emph{height} among all Tr\'emaux trees of~$G$.

\smallskip\noindent\textbf{Vertex Cover Size.}
A \emph{vertex cover} of a graph~$G=(V,E)$ is a set~$S\subseteq V$ such that for every~$\{u,v\}\in E$, we have~$u\in S$ or~$v\in S$. The \emph{vertex cover size} of~$G$ is the smallest~$\Card{S}$ over all vertex covers~$S$ of~$G$.

\begin{example}
Recall~$\Pi_1$ from Example~\ref{ex:running1}.
Figure~\ref{fig:graph-td} (left) depicts a PD of~$\mathcal{G}_{\Pi_1}$, whose width is~$2$, corresponding to the pathwidth, treewidth, and bandwidth of the graph.
The cutwidth of~$\mathcal{G}_{\Pi_1}$ is~$3$ and its cliquewidth is~$2$, as two labels are sufficient to construct~$\mathcal{G}_{\Pi_1}$.
The smallest FVSs of~$\mathcal{G}_{\Pi_1}$ are~$\{a\}$ and~$\{c\}$; the smallest sparse FVS is~$S_1=\{a,c\}$. Further, $S_1$ is the smallest vertex cover of~$\mathcal{G}_{\Pi_1}$.
Figure~\ref{fig:graph-td} (right) shows a Tr\'emaux tree of~$\mathcal{G}_{\Pi_1}$ of height~$2$ (treedepth of~$\mathcal{G}_{\Pi_1}$).
\end{example}

\begin{figure}[t]%
  \centering
  \shortversion{ %
    \includegraphics{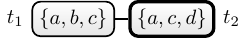}\qquad\quad
	\begin{tikzpicture}[node distance=7mm,every node/.style={circle,inner sep=2pt}]
\node (b2) [fill,right=11.5em of a,label={[text height=1.5ex]left:$a$}] {};
\node (a2) [fill,right=of b2,label={[text height=1.5ex,yshift=-0.15cm,xshift=-0.05cm]right:$c$}] {};
\node (c2) [fill,right=of a2,label={[text height=1.5ex,yshift=0.00cm,xshift=-0.05cm]right:$b$}] {};
\node (d2) [fill,below of=a2,label={[text height=1.5ex,yshift=0.09cm,xshift=-0.05cm]right:$d$}] {};
\draw[->] (b2) to (a2);
\draw [->] (a2) to  (c2);
\draw [->] (a2) to (d2);
\end{tikzpicture}
    \caption{(Left): A TD (PD)~$\mathcal{T}$ of~$\mathcal{G}_{\Pi_1}$ of Figure~\ref{fig:graphs}. (Right): A Tr\'emaux tree of~$\mathcal{G}_{\Pi_1}$.}
  }%
  \longversion{%
    \begin{subfigure}[c]{0.47\textwidth}
      \centering%
      \begin{tikzpicture}[node distance=7mm,every node/.style={fill,circle,inner sep=2pt}]
\node (a) [label={[text height=1.5ex,yshift=0.0cm,xshift=0.05cm]left:$d$}] {};
\node (e) [below of=a, label={[text height=1.5ex,xshift=-.34em,yshift=.42em]right:$a$}] {};
\node (d) [right of=e,label={[text height=1.5ex,yshift=0.09cm,xshift=-0.07cm]right:$b$}] {};
\node (c) [left of=e,label={[text height=1.5ex,yshift=0.09cm,xshift=0.05cm]left:$c$}] {};
\draw (a) to (c);
\draw (c) to (e);
\draw (d) to (e);
\draw (e) to (a);
\end{tikzpicture}%
      \input{graph0/td.tex}%
      \caption{Graph~$G_1$ and a tree decomposition of~$G_1$.}
      \label{fig:graph-td}
    \end{subfigure}
    \begin{subfigure}[c]{0.5\textwidth}
      \centering \begin{tikzpicture}[node distance=7mm,every node/.style={fill,circle,inner sep=2pt}]
\node (a) [label={[text height=1.5ex,yshift=0.0cm,xshift=0.12cm]left:$d$}] {};
\node (b) [right = 0.5cm of a,label={[text height=1.5ex,xshift=0.12cm]left:$a$}] {};
\node (c) [right = 0.5cm of b,label={[text height=1.5ex,xshift=0.05cm]left:$b$}] {};
\node (d) [right = 0.5cm of c,label={[text height=1.5ex,xshift=-0.1cm]right:$c$}] {};
\node (e) [right = 0.5cm of d,label={[text height=1.5ex,xshift=-0.1cm]right:$e$}] {};
\node (r3) [below = 0.5cm of a,label={[text height=1.5ex,xshift=-0.1cm]right:${r_3}$}] {};
\node (r1) [below = 0.5cm of b,label={[text height=1.5ex,xshift=-0.05cm]right:${r_1}$}] {};
\node (r2) [below = 0.5cm of c,label={[text height=1.5ex,xshift=-0.12cm]right:${r_2}$}] {};
\node (r4) [below = 0.5cm of d,label={[text height=1.5ex,xshift=-0.05cm]right:${r_4}$}] {};
\draw (a) to (r3);
\draw (c) to (r3);
\draw (b) to (r1);
\draw (c) to (r1);
\draw (b) to (r2);
\draw (c) to (r2);
\draw (d) to (r2);
\draw (d) to (r4);
\draw (e) to (r4);
\end{tikzpicture}%
      \begin{tikzpicture}[node distance=0.5mm]
\tikzset{every path/.style=thick}

\node (leaf0) [tdnode,label={[]left:$t_1$}] {$\{a,b, {r_1}, {r_2}\}$};
\node (leaf1) [tdnode,label={[xshift=0em]left:$t_2$}, above = 0.2cm of leaf0] {$\{b, d,{r_2}, {r_3}\}$};
\node (leaf2) [tdnode,label={[xshift=0em]right:$t_3$}, right = 0.2cm of leaf0]  {$\{c, e, {r_4}\}$};
\coordinate (middle) at ($ (leaf1.north east)!.5!(leaf2.north west) $);
\node (join) [tdnode,ultra thick,label={[xshift=-0.25em]right:$t_4$}, right = 0.25cm of leaf1] {$\{c, r_2\}$}; 

\draw [->] (leaf0) to (leaf1);
\draw [<-] (join) to (leaf1);
\draw [<-] (join) to (leaf2);
\end{tikzpicture}%
      \caption{Graph~$G_2$ and a tree decomposition of~$G_2$.}
      \label{fig:graph-td2.tex}%
    \end{subfqigure}
    \caption{Graphs~$G_1, G_2$ and two corresponding tree
      decompositions.}
  }%
  \label{fig:graph-td}%
\end{figure}

\futuresketch{
\begin{example}
\label{ex:running1}\label{ex:running}
Consider the following program
$\prog\eqdef$
%
$\SB\overbrace{ a \lor b \hsep}^{r_1};\, %
\overbrace{c \lor e \hsep d}^{r_2};\, %
\overbrace{d \lor e \hsep b}^{r_3};\, %
\overbrace{b \hsep e, \neg d}^{r_4};\, %
\overbrace{d \hsep \neg b}^{r_5} %
\SE$.
%
%
Observe that $\prog$ is head-cycle-free.
Then, $I\eqdef\{b, c, d\}$ is an answer set of~$\prog$,
since~$I\models\Pi$, and we can prove with ordering
$\varphi \eqdef\{b\mapsto 0, d\mapsto 1, c\mapsto 2\}$
atom~$b$ by rule~$r_1$, 
atom~$d$ by rule~$r_3$, and
atom~$c$ by rule~$r_2$.
Further answer sets are $\{b,e\}$,
$\{a,c,d\}$, and~$\{a,d,e\}$.


\end{example}}%
%


\futuresketch{
The following result for QBFs is known, where
it turns out that deciding $\QBFSAT$ remains $\ell$-fold exponential in the treewidth
of the primal graph (even when restricting the graph 
to the variables of the inner-most quantifier block).

\begin{proposition}[\cite{FichteHecherPfandler20}]\label{prop:lb}
Given a QBF~$Q=\exists V_1. \forall V_2, \ldots, \forall V_\ell. F$ of quantifier depth~$\ell$.
Then, unless ETH fails, the validity of~$Q$ cannot be decided
in time~$\tower(\ell, o(k))\cdot\poly(\var(Q))$, where~$k$
is the treewidth of the primal graph (even when restricted to vertices in~$V_\ell$).
\end{proposition}}

\futuresketch{
\subsection*{Reduction of SAT to normal ASP}

Note that~$w!/w^w=e^{-\mathcal{O}(k)}$ via Sterling's formula~\cite{LokshtanovMarxSaurabh11}.

Given a SAT formula~$F$ and a tree decomposition~$\mathcal{T}=(T,\chi)$ of the primal graph of~$F$ such that~$T=(N,A,n)$.
We reduce~$F$ to a normal program~$\Pi$ such that
$F$ is satisifiable if and only if~$\Pi$ has an answer set.
The reduction itself is guided by~$\mathcal{T}$ and the width of~$\mathcal{T}'$ is 
$\bigO{w/\log(w)}=\bigO{w/\log(w\cdot\log(w))}=\bigO{w/[\log(w)+\log(\log(w))]}$, 
where~$w$ is the width of~$\mathcal{T}$.
In total, with ASP, we can assign up to~$2^{\bigO{s\cdot\log(s)}}$ many states
for each bag~$\chi(t')$ of~$\mathcal{T}'$ of cardinality at most~$s$.
Concretely, we have exactly~$\Sigma_{I \subseteq \chi(t')} \Card{I}!$ many states.
We will use these states to simulate SAT, but thereby ``save'' in the treewidth.

Given any two bags~$\chi(t_1),\chi(t_2)$ of~$\mathcal{T}$ such that~$t_1$ is a children of~$t_2$, we construct the following ASP rules in~$\Pi$.
We assume an arbitrary total ordering~$\prec_{t_1}$ of the assignments~$I\in 2^{\chi(t_1)}$
in bag~$t_1$.
Further, we assume an arbitrary total ordering~$\prec_{t'_1}$ among $2^{\chi(t'_1)} \times ord(\chi(t'_1))$.
We use a mapping~$m: 2^{\chi(t_1)} \rightarrow [2^{\chi(t'_1)} \times ord(\chi(t'_1))]$,
which is a bijection between an assignment~$I$ in~$\chi(t_1)$ and an assignment~$I'$ over~$\chi(t'_1)$ together with
an ordering among those atoms in~$I'$.
The mapping~$m$ is naturally defined by assigning the elements in~$\prec_{t_1}$ to~$\prec_{t'_1}$ according to
the ordinal of the element.
Then, we add for each element~$I\in2^{\chi(t_1)}$ the rules~$0 { a_I } 1.$, but ensure an ``at most one'' behavior.
This can be done later on top of the TD. Further, we add~$v_j \leftarrow v_i, a_I.$
For each element~$I\in2^{\chi(t_1)}$ we add a parent~$t'_{1.i}$ to~$t'_1$.

\subsection*{Reduction of normal ASP to PASP using trick of exponential compression as used in the hierarchy}

TBD: Just reduce normal ASP to InvPASP problem, using the same ideas as in the hierarchy, exponential compression and such, then easily reduce InvPASP to PASP.
We have now a lower bound for PASP using normal ASP.
}

\section*{Vertex Cover Limits the Cost of Disjunction}\label{sec:vc}
In \cite{LampisMitsou17}, the authors showed that satisfiability of quantified Boolean formulas (QBFs) $\phi$ is possible in 
single exponential time in the size of the smallest vertex cover of the primal graph of $\phi$, if the size of clauses is bounded by a constant independent of $\phi$.

This begs the question of whether the same is possible for \ASP. The first idea that comes to mind, is to translate programs $\Pi$ to QBFs $\phi(\Pi)$. However, using the standard translation \cite{EglyEtAl00}, the smallest vertex cover of the program $\phi(\Pi)$ has at least size $|\at(\Pi)|$, and thus, cannot be bounded in the size of the smallest vertex cover~for~$\Pi$.

We show something stronger instead, namely, for programs with bounded rule size~$c$, e.g., $c=3$ for normalized programs, we give a polynomial kernel in terms of vertex cover size.

\begin{theorem}[Polynomial VC-Kernel]
\label{thm:primal_vc_kernel}
Let $\Pi$ be a program such that for every rule $r \in \Pi$ we have $|H_r| + |B_r^-| + |B_r^+| \leq c$ for~$c \in \mathbb{N}$. Further,  let $S \subseteq \at(\Pi)$ be a vertex cover of $\mathcal{G}_{\Pi}$.
Then there is a program $\Pi'$ where we have (i) $|\at(\Pi')| \leq$ $4^{c}\binom{|S|}{c}$ and (ii) $\Pi$ is consistent iff $\Pi'$ is consistent.
\end{theorem}
\begin{proof}[Proof (Sketch)]
The \emph{role} of an atom $a \not\in S$ in the program is defined by the rules where it occurs in. Since $S$ is a vertex cover of $\mathcal{G}_{\Pi}$, all other atoms in the rules that contain $a$ must be in $S$. Because the rule size is bounded by $c$, we can bound the number of different roles. However, we can show that if two atoms have the same role, then we can remove one of them without changing consistency of the program.
\end{proof}

A similar result holds if the primal graph does not directly represent the structure of \emph{type 1} and \emph{type 2} rules of the form~$a \leftarrow \neg b$ or~$a \vee b\leftarrow$, respectively. This is important, as these rules are oftentimes required to model ``guesses'' or use the full power of \ASP semantics, respectively. Indeed, this will otherwise be the source of a large vertex cover size.
To this end, let the $\mathcal{G}_\Pi^i$, for $i = 0, 1, 2$ be the primal graphs consisting of the type $i$ rules of a program~$\Pi$, where type 0 rules are rules that are not of type 1 or 2.

\begin{corollary}\label{cor:vcext}
Let $\Pi$ be a program such that for every rule $r \in \Pi$ we have $|H_r| + |B_r^-| + |B_r^+| \leq c$ for~$c \in \mathbb{N}$, where every atom has at most one neighbor in $\mathcal{G}_{\Pi}^1$ or $\mathcal{G}_{\Pi}^2$ (that is not a neighbor in $\mathcal{G}_{\Pi}^0$). Furthermore, let $S \subseteq \at(\Pi)$ be a vertex cover of $\mathcal{I}_{\Pi}^0$.
Then there exists a program $\Pi'$ such that (i) $|\at(\Pi')| \leq 3 \cdot 4 \cdot (4^{c}\binom{|S|}{c})^2 + 4^{c}\binom{|S|}{c}$ and (ii) $\Pi$ is consistent iff $\Pi'$ is consistent.
\end{corollary}
%
%

The algorithm above also works for the incidence graph.
\begin{corollary}\label{cor:vcinc}
Let $\Pi$ be a program such that for every rule $r \in \Pi$ we have $|H_r| + |B_r^-| + |B_r^+| \leq c$ for~$c \in \mathbb{N}$ and let $S \subseteq \at(\Pi)$ be a vertex cover of $\mathcal{I}_{\Pi}$.
Then there exists a program $\Pi'$ such that (i) $|\at(\Pi')| \leq 4^{c}\binom{|S|}{c}c^c$ and (ii) $\Pi$ is consistent iff $\Pi'$ is consistent. 
\end{corollary}

The polynomial kernels above immediately give rise to the following runtime result.

\begin{corollary}\label{cor:vcruntime}
Let $\Pi$ be a program such that for every rule $r \in \Pi$ we have $|H_r| + |B_r^-| + |B_r^+| \leq c$ for~$c \in \mathbb{N}$ and vertex cover size $k$ of the primal graph.
Then we can decide consistency of $\Pi$ in time $\mathcal{O}(2^{2\cdot4^{c}\binom{k}{c}}\poly(|\at(\Pi)|))$.
\end{corollary}

\section*{Decrease Cyclicity and Treedepth by the Power of Disjunction}
\label{sec:fvs}
\begin{figure*}[t]
{
\begin{flalign}
	&\textbf{Guess Interpretation, Pointers, and Values}\hspace{-10em}\notag\\
	\label{red:guess}&x\vee \overline{x}\leftarrow \qquad sat_r\vee \overline{sat_r}\leftarrow\qquad b_{j}^i \vee \overline{b_{j}^i}\leftarrow\qquad v_{j}\vee \overline{v_{j}}\leftarrow\hspace{-10em}  & \text{for every }x{\,\in\,}\at(\Pi), r{\,\in\,} \Pi, 0\leq i< \ceil{\log(\Card{S})}, 1\leq j \leq 3 
	\\
%
%
%
%
	&\textbf{Saturate Pointers and Values}\hspace{-10em}\notag\\
	\label{red:saturate}&b_{j}^i \leftarrow sat\quad\qquad \overline{b_{j}^i} \leftarrow sat \qquad\quad v_{j} \leftarrow sat\quad\qquad \overline{v_{j}} \leftarrow sat\hspace{-4em}  &\text{for every }0\leq i< \ceil{\log(\Card{S})}, 1\leq j\leq 3\\
	%
%
	%
	%
	%
	%
	\label{red:keepsat}&sat_r \leftarrow sat\qquad \overline{sat_r}\leftarrow sat\qquad \leftarrow \neg sat &\text{for every }r\in\Pi\\
	%
	%
	 %
%
%
%
	&\textbf{Synchronize Pointers and Values}\hspace{-10em}\notag\\
%
%
%
%
%
%
\label{red:satptr}&sat \leftarrow sat_r, \dot{b} &\text{for every }r\in\Pi, 1\leq j\leq 3, x\in S\cap\at(r), j=\ord(r,x), b\in\bval{x}{j}\\
\label{red:satguess}&sat \leftarrow   b^0, \ldots, b^n, \overline{v_{j}}, x\qquad sat \leftarrow b^0, \ldots, b^n, v_{j}, \overline{x}\hspace{-10em} &\text{for every }1\leq j\leq 3, x\in S, \bval{x}{j}=\{b^0, \ldots, b^n\}\\
%
%
%
%
%
%
	&\textbf{Check Satisfiability of Rules}\hspace{-10em}\notag\\
	&sat \leftarrow sat_r, v_{j}  & \text{ for every }r\in\Pi,x\in S\cap (H_r\cup B_r^-), 
	\label{red:head-true}j=\ord(r,x)\\
	&sat \leftarrow sat_r, \overline{v_{j}}  & \text{ for every }r\in\Pi,x\in S\cap B_r^+, \label{red:bodyplus-false}j=\ord(r,x)\\
	%
	%
	%
%
&sat\leftarrow sat_r, x  & \text{ for every }r\in\Pi,x\in (H_r\cup B_r^-) \setminus S 
	 \label{red:bodyneg-false1} \\
%
%
&sat\leftarrow sat_r, \overline{x}  & \text{ for every }r\in\Pi,x\in  B_r^+\setminus S 
	 \label{red:bodyneg-false2} \\
%
	%
	\label{red:sats} &sat \leftarrow \overline{sat_{r_1}}, \ldots \overline{sat_{r_n}}&\text{where }\Pi=\{r_1,\ldots,r_n\}
\end{flalign}
}\caption{The reduction~$\mathcal{R}_{fvs}$ that takes a normalized fully tight program~$\Pi$ and a sparse feedback vertex set~$S$ of~$\mathcal{G}_{\Pi}$.}
\label{fig:red}
\end{figure*}

\label{sec:main}
In this section
we show how to translate a (tight) logic program
to a disjunctive program,
thereby exponentially decreasing the program's cyclicity, i.e., it's feedback vertex size and explicitly utilizing the structural power of disjunctive \ASP.
%
%
The reduction uses head-cycles and disjunction in order to carry out the exponential parameter decrease, which will yield new lower bounds for disjunctive \ASP.
%
In fact, we even prove a stronger lower bound, which already holds
for (i) normalized fully tight programs and (ii) sparse FVS size.
\futuresketch{
We proceed in two steps. First, we translate the Boolean formula into a formula,
where it is guaranteed that there exists a nice tree decomposition of the formula's primal graph
such that every variable of the formula appears only in constantly many bags.

\subsection*{Elimination of Decomposition-Global Variables}
In the first step we show how one can eliminate decomposition-global variables
of a Boolean formula~$F$, i.e., we reduce to a new formula~$F'$,
where variables are only in at most three decomposition bags.

Let therefore
$\mathcal{T}=(T,\chi)$ be a nice tree decomposition of~$\mathcal{G}_{F}$. 
The idea is as follows: Every variable~$v$ in a bag~$\chi(t)$ for a node~$t$ of~$T$
is bound to~$t$, i.e., we require a new variable~$v_t$.
So the new formula~$F'$ has as variables the fresh set~$\{v_t \mid t\text{ in }T, v\in \chi(t)\}$
of variables. Then, for every node~$t$ of~$T$ with~$t'\in \children(t)$ and~$x\in \chi(t)\cap\chi(t')$,
we ensure equivalence: \inlineequation[botup:prop]{x_t \longleftrightarrow x_{t'}}. 
Finally, for every node~$t$ of~$T$ and clause~$(l_1 \vee \ldots \vee l_o)\in \{c \mid c\in F, \var(c)\subseteq \chi(t) \}$, we take care that the clause is satisfied in~$t$: \inlineequation[botup:clause]{l_1^t \vee \ldots \vee l_o^t}, 
where function~$\cdot^t$ takes a literal over some variable~$x$ and replaces the occurrence of~$x$ by~$x_t$.

Observe that indeed the resulting formula~$F'$ is satisfiable if and only if~$F$ is satisfiable and there is even a bijective correspondence between models of~$F$ and models of~$F'$. Moreover, we can easily construct a tree decomposition~$\mathcal{T}'\eqdef (T,\chi')$ of~$\mathcal{G}_{F'}$ such that every variable of~$F'$ occurs in at most three bags of~$\mathcal{T}'$. Thereby, the tree~$T$ of~$\mathcal{T}'$  is used as before, we only need to define~$\chi'$, which we give for every node~$t$ of~$T$ as follows: $\chi'(t)\eqdef \{x_t \mid x\in\chi(t)\} \cup \{x_{t^*} \mid \chi(t)\neq\emptyset,x\in \chi(t^*), t^* \text{ is the parent of }t\text{ in }T\}$.
The construction of~$\mathcal{T}'$ ensures that~$\mathcal{T}'$ is a TD of~$\mathcal{G}_{F'}$. Further, the width of~$\mathcal{T}'$ is at most doubled and every variable of~$F'$ appears in at most three different bags of~$\mathcal{T}'$. 

\smallskip
In the following, we \emph{only} assume such a formula~$F'$ and a nice TD~$\mathcal{T}''$ of~$\mathcal{G}_{F'}$, which is obtained from~$\mathcal{T}'$ above, by adding an auxiliary node~$t'$ for every node~$t$ that has two child nodes and making~$t$ the parent of~$t'$, i.e., $t'$ serves as a fresh join node that is placed between~$t$ and their child nodes.  
As a result, every variable of~$F'$ then appears in at most four bags; we refer to~$F'$ as \emph{local formula}
and~$\mathcal{T}''$ as \emph{join-local tree decomposition (of~$\mathcal{G}_F$)}.
This first step is a preprocessing step, providing a normal form of formulas and TDs, making it easier to present and grasp the second part of our~reduction. 
}



The idea of this reduction~$\mathcal{R}_{fvs}$ is as follows.
We take a normalized fully tight program~$\Pi$ and a sparse feedback vertex set~$S$ of~$\mathcal{G}_{\Pi}$ of size~$\Card{S}=k$.
From this we construct a disjunctive program, where 
we decouple the atoms of~$\Pi$ and we model 
three pointers to~$S$ to check satisfiability of~$\Pi$.
Since~$\Pi$ is normalized, i.e., ever rule contains at most three atoms, indeed three pointers are sufficient to refer to all the atoms of a rule.
These three pointers can be expressed by using~$3\log(k)$ many atoms,
which will allow us to achieve the desired compression of feedback vertex size from~$k$ to~$\log(k)$.

This reduction~$\mathcal{R}_{fvs}$ constructs a disjunctive program,
thereby relying on the atoms~$\at(\Pi)$ of the given (normalized) fully tight program~$\Pi$ and the following additional auxiliary atoms.
We use atoms~$sat$ to indicate
satisfaction for the interpretation over~$\at(\Pi)$ of the program.
Further, for every~$r\in \Pi$, if $sat_r$ holds, satisfaction of~$r$ shall be checked.
Further, we require bit atoms of the form~$b_j^i$ for three pointers ($1\leq j\leq 3$), which indicates that the $i$-th bit of pointer~$j$ is set to one, and we need~$\ceil{{\log(\Card{S})}}$ many of these bits, i.e., $0\leq i < \ceil{{\log(\Card{S})}}$.
In order to also unset these atoms (i.e., set the bits to zero), we require corresponding inverse copies of the form~$\overline{b_j^i}$,
with the intended meaning that the $i$-th bit of pointer~$j$ is set to zero.
Intuitively, a subset over these atoms of cardinality $\ceil{{\log(\Card{S})}}$ for a fixed~$j$ addresses exactly one atom that is contained in~$S$. 
We briefly need to define auxiliary notation. For every atom~$x\in S$, we let~$\bval{x}{j}$ be the set of these bit atoms for pointer~$j$
of cardinality $\ceil{{\log(\Card{S})}}$ that uniquely addresses~$x$.
To this end, one may assume that the elements in~$S$ are given in any arbitrary, but fixed order and that~$\bval{x}{j}$ then binary-encodes the ordinal position of~$x$ in~$S$. 
Finally, in addition to the capability of referring to atoms in~$S$, we also need to set the value for the target of the three pointers, resulting in atoms~$v_1$, $v_2$, and $v_3$.
%

\smallskip
\noindent \textbf{The Reduction.} We are ready to proceed with the formal description of our reduction,
as given in Figure~\ref{fig:red}\footnote{For brevity, $\dot{b}$ for atom~$b$ refers to~$\overline{b_j^i}$ if~$b{=}b_j^i$ and to $b_j^i$ if $b{=}\overline{b_j^i}$.}. 
To this end, we take our normalized fully tight 
program~$\Pi$ and the feedback vertex set~$S$ of~$\mathcal{G}_{\Pi}$ and construct a program~$\Pi'$ according to~$\mathcal{R}_{fvs}$ as follows.
For improved readability, reduction~$\mathcal{R}_{fvs}$ is split into four different groups.
In the first group, consisting of Rules~(\ref{red:guess}),
 we guess the interpretation, the bits that will form pointers, the pointer target's values, as well as the rules of~$\Pi$ that shall be checked.
The atoms appearing under disjunction will be subject to saturation, so
any answer set can only contain all of these atoms, which is ensured by Rules~(\ref{red:saturate})--(\ref{red:keepsat})  of the second group, where~(\ref{red:keepsat}) requires $sat$ to be in every answer set.
So, whenever we check an answer set candidate~$M$, for \emph{every combination} of these atoms under disjunction, there must not exist a $\subseteq$-smaller model 
of~$\Pi^M$, i.e., intuitively, these atoms are implicitly universally quantified. 

The remainder boils down to synchronizing pointers and their values, which is achieved in the third group of rules, and checking rules in the fourth group.
Rules~(\ref{red:satptr}) ensure that whenever we need to check 
rule~$r$, i.e., in case~$sat_r$ holds, as soon as any bit of the three pointers is not in accordance with the bit combination addressing the three atoms of~$r$, we obtain~$sat$. In other words, as soon as we refer to the atoms of~$r$ incorrectly, we just skip those interpretations.
Similarly, by Rules~(\ref{red:satguess}), 
we have that whenever the $j$-pointer refers to an atom~$x$ that is contained in~$S$ and
the pointer value~$v_j$ is contained in the answer set candidate, but~$x$ is not (or vice versa), we also obtain~$sat$.
Note that it is crucial for the feedback vertex set of the resulting program that $sat_r$ does not appear in any of the two rules.

Finally, a rule~$r$ is satisfied if the pointer value~$v_j$ for the $j$-th atom is set accordingly, ensured by Rules~(\ref{red:head-true}) and~(\ref{red:bodyplus-false}), or if any of those rule atoms not in~$S$ are set accordingly, see Rules~(\ref{red:bodyneg-false1}) and~(\ref{red:bodyneg-false2}).
The only remaining rule is~(\ref{red:sats}), which is required to avoid $\subseteq$-smaller reduct models containing every atom of the form~$\overline{sat_r}$ and none of the form~$sat_r$.
Intuitively, this is essential, since not checking any of these rules~$r$ would lead to a $\subseteq$-smaller reduct model. 
%

Before we discuss the consequences of this rule in more details, and we briefly demonstrate the reduction below.

%

\begin{figure}[t]
\begin{tikzpicture}[node distance=7mm,every node/.style={circle,inner sep=2pt}]
\node (b) [fill,label={[text height=1.5ex,xshift=.0em]above:$b$}] {};
\node (y) [fill,below=2.35em of b,label={[text height=1.5ex,xshift=.0em]below:$r_2^b$}] {};
\node (z) [fill,below=2.35em of y,label={[text height=1.5ex,xshift=.0em]below:$d$}] {};
%
%
\draw (y) to (b);
%
\node (c) [tdnode,ellipse,draw=black,text height=5em,yshift=-2.5em,text width=2em,xshift=0.0em,color=white,right=1.2em of b,label={[text height=1.5ex,yshift=0.00cm,xshift=0.00cm]below:$S$}] {}; 
\node (s1) [draw=black,right=2.3em of b,yshift=-.0em,label={[text height=1.5ex,yshift=0.00cm,xshift=0.35em]right:$a$}] {};
\node (s2) [draw=black,below=2.35em of s1,label={[text height=1.5ex,yshift=0.00cm,xshift=0.35em]right:$c$}] {};
\node (s3) [draw=black,below=2.35em of s2,label={[text height=1.5ex,yshift=0.00cm,xshift=0.35em]right:$r_1^b$}] {};
%
\draw (s1) to (s2);
\draw [out=-2,in=2] (s1) to (s3);
%
%
\draw [dashed,in=20,out=140] (s1) to (b);
\draw [dashed,in=20,out=140] (s1) to (y);
\draw [dashed,in=20,out=140] (s1) to (z);
\draw [dashed,in=20,out=140] (s2) to (y);
\draw [dashed,in=-20,out=-140] (s3) to (y);
\draw [dashed,in=20,out=170] (s2) to (z);
\draw [dashed,in=20,out=140] (s2) to (b);
\draw [dashed,in=20,out=140] (s3) to (b);
%
%
%
%
%
\node (b2) [fill,right=7.0em of b,label={[text height=1.5ex,yshift=.2em,xshift=.35em]left:$b$}] {};
\node (x2s) [fill,right=1.15em of b2,yshift=0.4em,label={[text height=1.5ex,xshift=-.3em,yshift=1em]below:{\tiny $sat_1$}}] {};
\node (x2s2) [fill,right=-.1em of b2,yshift=1em,label={[text height=1.5ex,xshift=-.35em,yshift=.3em]right:{\tiny $sat_2$}}] {};
\node (y2) [fill,below=2.35em of b2,label={[text height=1.5ex,xshift=.4em,yshift=.15em]left:{ $r_2^b$}}] {};
\node (y2sz) [fill,below=.35em of y2,label={[text height=1.5ex,xshift=.5em,yshift=-.15em]left:{\tiny $\overline{r_2^b}$}}] {};
\node (y2s3) [fill,below=.15em of b2,label={[text height=1.5ex,xshift=-.45em,yshift=.1em]right:{\tiny $\overline{b}$}}] {};
\node (y2s) [fill,below=1.2em of b2,label={[text height=1.5ex,xshift=.45em,yshift=.45em]left:{\tiny $sat_5$}}] {};
\node (z2) [fill,below=2.5em of y2,label={[text height=1.5ex,xshift=.35em]left:$d$}] {};
\node (z2x) [fill,below=.25em of z2,label={[text height=1.5ex,yshift=1.4em,xshift=.5em]below:{\tiny $\overline{d}$}}] {};
\node (z2s) [fill,right=.5em of y2,yshift=.1em,label={[text height=1.5ex,xshift=.3em,yshift=-.75em]above:{\tiny $sat_7$}}] {};
\node (z2s2) [fill,below right=.9em of y2,yshift=.25em,label={[text height=1.5ex,xshift=-.25em,yshift=-.1em]right:{\tiny $sat_8$}}] {};
\node (f2s) [fill,below left=.75em of z2,label={[text height=1.5ex,xshift=.15em,yshift=1.2em]below:{\tiny$sat_{11}$}}] {};
%
\node (d2s2) [fill,above=.3em of z2,label={[text height=1.5ex,xshift=2.3em,yshift=.5em]left:{\tiny$sat_3$}}] {};
%
%
\node (sat9) [fill,right=0.5em of z2,yshift=.5em,label={[text height=1.5ex,xshift=-.35em,yshift=.15em]right:{\tiny $sat_4$}}] {};
\node (e2s) [fill,below right=.9em of z2,label={[text height=1.5ex,xshift=0.25em,yshift=.9em]below:{\tiny$sat_{9}$}}] {};
\draw (b2) to (x2s2);
\draw (b2) to (x2s);
\draw (y2) to (b2);
\draw (z2x) to (e2s);
\draw (z2x) to (f2s);
\draw (z2) to (z2x);
\draw (z2) to (d2s2);
\draw (z2) to (sat9);
\draw (y2sz) to (y2);
\draw (y2sz) to (z2s);
\draw (y2sz) to (z2s2);
%

\node (sx) [draw=white,right=9em of b2,yshift=1em,label={[text height=1.5ex,yshift=0.00cm,xshift=0.35em]right:$ $}] {};

\node (c2) [tdnode,ellipse,draw=black,text height=5em,yshift=-2.6em,text width=7em,xshift=-.95em,color=white,right=4.35em of b2,label={[text height=1.5ex,yshift=0.00cm,xshift=0.00cm]below:$S'$}] {}; 
\node (sat6) [fill,above=-.45em of sx,xshift=.45em,yshift=.15em,label={[text height=1.5ex,xshift=-.35em,yshift=.17em]right:{\tiny $sat_6$}}] {};
\node (d2s) [fill,left=.2em of sat6,label={[text height=1.5ex,xshift=.45em,yshift=-.12em]left:{\tiny$sat_{10}$}}] {};
\node (s1z) [fill=black,right=18.5em of s1,label={[text height=1.5ex,yshift=0.00cm,xshift=-0.0em]above:$a$}] {};
\node (s2z) [fill=black,below=2.35em of s1z,label={[text height=1.5ex,yshift=0.00cm,xshift=-0.0em]above:$c$}] {};
\node (s3z) [fill=black,below=2.35em of s2z,label={[text height=1.5ex,yshift=0.00cm,xshift=-0.0em]above:$r_1^b$}] {};

\node (s12) [draw=black,right=5.75em of b2,yshift=+.0em,label={[text height=1.5ex,yshift=0.00cm,xshift=-0.35em]right:$b_1^0$}] {};
\node (s13) [draw=black,right=10.5em of b2,yshift=-.0em,label={[text height=1.5ex,yshift=0.00cm,xshift=-0.3em]right:$\overline{b_3^0}$}] {};
\node (s22) [draw=black,below=2.35em of s12,label={[text height=1.5ex,yshift=0.00cm,xshift=-0.25em]right:$b_1^1$\; $\ldots$}] {};
\node (s23) [draw=black,below=2.35em of s13,label={[text height=1.5ex,yshift=0.00cm,xshift=-0.3em]right:$\overline{b_3^1}$}] {};
\node (vs22) [draw=black,below=2.35em of s22,label={[text height=1.5ex,yshift=0.00cm,xshift=-0.35em]right:$v_1$}] {};
\node (vsdots) [draw=black,below=2.35em of s22,label={[text height=1.5ex,yshift=0.00cm,xshift=0.35em]right:$ $}] {};
\node (vsdots2) [draw=black,below=2.35em of s23,label={[text height=1.5ex,yshift=0.00cm,xshift=-0.3em]right:$\overline{v_3}$}] {};
\node (sat) [draw=black,below=2.35em of s22,xshift=2.1em,label={[text height=1.5ex,yshift=0.00cm,xshift=-0.35em]right:$sat$}] {};
%
%
%
\draw [dashed,in=20,out=140] (s23) to (sx);
\draw [dashed,in=20,out=140] (vsdots2) to (sx);
\draw [dashed,in=-80,out=180] (s13) to (d2s);
\draw [dashed,in=-80,out=180] (s23) to (d2s);
\draw [dashed,in=-80,out=150] (vsdots2) to (d2s);
%
%
%
%
\draw [dashed,in=15,out=140] (s12) to (y2s);
\draw [dashed,in=15,out=140] (s22) to (y2s);
\draw [dashed,in=15,out=140] (vs22) to (y2s);
\draw [dashed,in=15,out=140] (s12) to (y2sz);
\draw [dashed,in=15,out=140] (s22) to (y2sz);
\draw [dashed,in=15,out=140] (vs22) to (y2sz);
\draw [dashed,in=15,out=140] (s12) to (b2);
\draw [dashed,in=25,out=140] (s22) to (b2);
\draw [dashed,in=35,out=140] (vs22) to (b2);
\draw [dashed,in=-35,out=200] (s1z) to (c2);
\draw [dashed,in=-35,out=200] (s2z) to (c2);
\draw [dashed,in=-35,out=200] (s3z) to (c2);
%
%
\draw [dashed,in=-7,out=260] (s12) to (z2x);
\draw [dashed,in=-15,out=260] (s22) to (z2x);
\draw [dashed,in=-23,out=260] (vs22) to (z2x);
%
%
%
%
%
%
%
%
%
\end{tikzpicture}
\caption{Visualization of the structure 
of~$\mathcal{R}_{fvs}$ for some normalized fully tight program~$\Pi$. (Left): Primal graph~$\mathcal{G}_{\Pi}$ together with a sparse FVS~$S$ that connects~$\mathcal{G}_{\Pi}$. (Right): $\mathcal{G}_{\Pi'}$ (simplified) and a sparse FVS~$S'$ of~$\Pi'$, obtained by~$\mathcal{R}_{fvs}$. 
}\label{fig:sketch}
\end{figure}
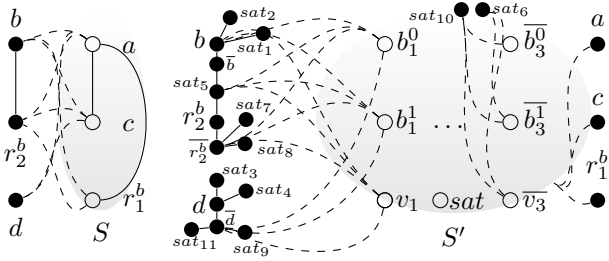

\begin{example}
Recall program~$\Pi_2$ over~$6$ atoms
and~$11$ rules, and~$S_1$ from Example~\ref{ex:running1}. Observe that, e.g., $S_2=S_1\cup\{r_1^b\}$ is a sparse FVS of~$\mathcal{G}_{\Pi_2}$.
Below we list selected rules of~$\mathcal{R}_{fvs}(\Pi_2, S_2)$.
Since~$\Card{S_2}\leq 3$, 
we require~$3\cdot \ceil{\log(\Card{S_2})}=6$ bit atoms~$b_j^0, b_j^1$ and $3$ values~$v_j$ ($1\leq j\leq 3$), to simultaneously address and assign up to~$3$ atoms in a rule of~$\Pi_2$.

Assuming~$\ord(r_4,a)=1$ and~$\ord(r_4,c)=2$,
as well as~$\bval{a}{1}=\{\neg b_1^0, \neg b_1^1\}$ and~$\bval{c}{2}=\{\neg b_2^0, b_2^1\}$, 
we construct these rules for~$r_4$: 
Besides Rules~(\ref{red:guess})--(\ref{red:keepsat}) and~(\ref{red:sats}),
Rules~(\ref{red:satptr}) ensure that if~$sat_{r_4}$ holds, we address~$a$ and~$c$ by~$sat \leftarrow sat_{r_4},  b_1^0$; $sat \leftarrow sat_{r_4},  b_1^1$; $sat \leftarrow sat_{r_4}, b_2^0$; and $sat \leftarrow sat_{r_4}, \neg b_2^1$.
Pointer values are in sync with atoms in~$S$ 
by Rules~(\ref{red:satguess}).
Rules~(\ref{red:head-true})--(\ref{red:bodyneg-false2}) ensure satisfiability of~$r_4$ by $sat \leftarrow sat_{r_4}, \neg v_1$;  $sat \leftarrow sat_{r_4}, v_2$; and  $sat \leftarrow sat_{r_4}, d$.

Figure~\ref{fig:sketch} visualizes the relation between a primal graph~$\mathcal{G}_{\Pi}$ and the resulting primal graph $\mathcal{G}_{\mathcal{R}_{fvs}(\Pi, S)}$,
along the lines of our running example where $\Pi=\Pi_2$ and $S=S_2$. 
\end{example}

\smallskip
\noindent
\textbf{Results.} 
 First, we show runtime results and correctness. 

\begin{lemma}[Runtime \& Correctness]\label{lem:rtmcorr}
Let~$\Pi$ be any normalized fully tight program, $S$ be any sparse FVS of~$\mathcal{G}_{\Pi}$, and let~$\Pi'{=}\mathcal{R}_{fvs}(\Pi, S)$. $\mathcal{R}_{fvs}$ runs in time~$\mathcal{O}(\Card{\at(\Pi)}+\Card{\Pi}\cdot\log(\Card{S})+\Card{S})$ and it is correct: Any answer set~$M$ of~$\Pi$ can be extended to an answer set~$M'$ of~$\Pi'$. Further, for any answer set~$M'$ of~$\Pi'$, $M'\cap\at(\Pi)$ is an answer set of~$\Pi$.
\end{lemma}

Leaving Rule~(\ref{red:sats}) aside, we can show that the reduction significantly decreases the feedback vertex size.
Interestingly, while the disjunction over atoms~$\at(\Pi)$ in Rules~(\ref{red:guess}) is not required for saturation, it serves in preserving sparsity. 

\begin{theorem}\label{thm:compr}
Let~$\Pi$ be any normalized fully tight program, $S$ be any sparse FVS of~$\mathcal{G}_{\Pi}$, and let~$\Pi'=\mathcal{R}_{fvs}(\Pi, S)$. Then, there is a sparse FVS of
$\mathcal{G}_{\Pi''}$ of size~$\mathcal{O}(\ceil{\log(\Card{S})})$, where~$\Pi''$ is a normalized program of~$\Pi'\setminus\{r\}$, with $r$ being Rule~(\ref{red:sats}).
\end{theorem}

We do not expect that this result can be improved without excluding Rule~(\ref{red:sats}).
Intuitively, the problem is that normalizing this rule significantly incrases 
the feedback vertex size.
However, for a given sparse feedback vertex set~$S$ of the primal graph~$\mathcal{G}_{\Pi}$, we have a direct
relationship between~$\Card{S}$ and the feedback vertex size of the natural incidence graph~$\mathcal{I}_{\Pi'}$
of the resulting program~$\Pi'$.
We obtain the following result.

\begin{lemma}[Compression]\label{lem:compr:fs}
Let~$\Pi$ be a normalized fully tight program, $S$ be a sparse FVS of~$\mathcal{G}_{\Pi}$, and $\Pi'{=}\mathcal{R}_{fvs}(\Pi, S)$. Then, there is a sparse FVS of 
$\mathcal{I}_{\Pi'}$ of size $\mathcal{O}(\ceil{\log(\Card{S})})$.
\end{lemma}
\begin{proof}
We define~$S''\eqdef S' \cup\{r\}$, where~$S'$ is defined in Lemma~\ref{thm:compr} and~$r$ refers to Rule~(\ref{red:sats}), which will be a FVS of~$\mathcal{I}_{\Pi'}$ for the program~$\Pi'$. 
Obviously, $S''$ breaks-up the cycles in~$\mathcal{I}_{\Pi'}$ caused by Rule~(\ref{red:sats}).
Further,  since by Theorem~\ref{thm:compr}, $S'$ is a sparse FVS of~$\mathcal{G}_{\Pi'}$, we follow that removing~$S''$ from $\mathcal{I}_{\Pi'}$ results in an acyclic graph.
\end{proof}

Note that Rules~(\ref{red:satguess}) 
can be easily normalized without increasing the feedback vertex size. 
The result above allows us to directly prove the following lower bound. 
\begin{theorem}[FVS LB]\label{thm:lb:fvs}
Let~$\Pi$ be any disjunctive program. Under ETH the consistency of~$\Pi$ cannot be decided in time $2^{2^{o(k)}}\cdot\poly(\Card{\at(\Pi)})$ for (sparse) FVS size~$k$ of $\mathcal{I}_\Pi$.
\end{theorem}
\begin{proof}
Assume towards a contradiction that despite ETH, we can decide~$\Pi$ in time $2^{2^{o(k)}}\cdot\poly(\Card{\at(\Pi)})$. 
Then, we take any 3-CNF formula~$F$ and transform it into a normalized fully tight
program~$\Pi'{\,\eqdef\,} \{a \leftarrow \neg \overline{a}, \overline{a} \leftarrow \neg a \mid a\in\var(F)\}\allowbreak \cup \{\leftarrow \hat\ell_1, \hat\ell_2, \hat\ell_3 \mid (\ell_1 \vee \ell_2 \vee \ell_3)\in F\}$, where~$\hat \ell {\,\eqdef\,} \neg\ell$ if $\ell{\,\in\,}\var(F)$ and~$\hat\ell{\,\eqdef\,}\ell$ otherwise.
We compute a sparse FVS~$S$ of~$\mathcal{G}_{\Pi'}$ in time~$3.619^{\mathcal{O}(k)}\cdot\poly(\Card{\at(\Pi')})$~\cite{KociumakaPilipczuk14}.
We apply our reduction and~construct $\Pi{\eqdef}\mathcal{R}_{fvs}(\Pi', S)$, running in polynomial time. By Lemma~\ref{lem:compr:fs}, there is a FVS of~$\mathcal{I}_{\Pi}$ of size~$k{\eqdef}\mathcal{O}(k)$ with $k{\eqdef}\log(\Card{S})$.
Solving~$\Pi$ in time~$2^{2^{o(k)}}\cdot\poly(\Card{\at(\Pi')})$ implies solving $F$ in time~$2^{o(\Card{S})}\cdot\poly(\Card{\var(F)})$, which contradicts ETH.
\end{proof}

We get a similar result for the primal graph, assuming that just a single rule is not represented in this graph. 

The lower bound can be strengthened to the \emph{distance~$k$ to almost paths}, 
even if the path lengths are linear in $k$. This distance \emph{counts the number of vertices} that need to be removed from a graph, to obtain disconnected paths, where every vertex may have one additional degree-$1$ neighbor.
\begin{theorem}\label{thm:almostpath}
Let~$\Pi$ be any disjunctive program. Under ETH, the consistency of~$\Pi$ cannot be decided in time $2^{2^{o(k)}}\cdot\poly(\Card{\at(\Pi)})$ for distance~$k$ to almost paths of $\mathcal{I}_\Pi$. The result still holds if the largest path length is in $\mathcal{O}(k)$.
\end{theorem}

Even further, we also get a tight lower bound for treedepth.

\begin{theorem}[TDP LB]\label{lem:compr:tdp}
Let~$\Pi$ be any disjunctive program. Under ETH the consistency of~$\Pi$ cannot be decided in time $2^{2^{o(k)}}\cdot\poly(\Card{\at(\Pi)})$ for treedepth~$k$ of $\mathcal{I}_\Pi$.
\end{theorem}

\smallskip
\noindent\textbf%
{A Note on Normal Programs.}
Our reductions still work for non-tight programs,
assuming that besides Clark's completion~\cite{Clark77}, 
a suitable technique for treating positive cycles~\cite{LinZhao03}, 
e.g., a treewidth-aware~\cite{Hecher22} 
level ordering~\cite{Janhunen06}, 
and normalization, has been applied.
%
%
However, for  separating disjunctive from normal programs regarding structural hardness, normalized fully tight programs are aready sufficient.  

\section*{Hardness for Bandwidth and Cutwidth}\label{sec:bw}

The reduction of the previous section can be further generalized 
to work also for other parameters more general than treewidth.
To show double-exponential lower bounds for these parameters, 
we design a reduction on tree decompositions. Then we will obtain stronger 
results by tweaking the reduction and 
considering more restricted decompositions.

For the ease of presentation, we rely on the following TDs.
\begin{definition}[Annotated TD]
Let~$\Pi$ be a program, $G_\Pi$ be a graph representation of~$\Pi$,
and~$\mathcal{T}{\,=\,}(T,\chi)$ be a nice TD of $G_\Pi$ such that $T{\,=\,}(V,E)$.
Further, let~$\varphi: (\at(\Pi)\cup \Pi) \rightarrow V$  be an injective mapping
such that (i) for every~$a\in \at(\Pi)$,  we have $a\in\chi(\varphi(a))$
and (ii) for every~$r\in\Pi$, we have $\at(r) \subseteq \chi(\varphi(r))$.
Then, we call $(T,\chi,\varphi)$ an \emph{annotated TD}.
\end{definition}
Note that an annotated TD ensures that every atom and rule
get assigned a unique node, whose bag contains the atom and the rule's atoms,
respectively. Any TD can be annotated by
greedily assigning suitable atoms and rules to a node 
and, if necessary, duplicating this node to ensure injectivity.

\begin{figure*}[t]
{
\begin{flalign}
	&\textbf{Guess Interpretation, Pointers, and Values}\hspace{-10em}\notag\\
	\label{redt:guess}&x\vee\overline{x}\leftarrow\;\;\; sat_r\vee\overline{sat_r}\leftarrow \;\;\; b_{t,j}^i \vee \overline{b_{t,j}^i}\leftarrow \;\;\; v_{t,j}\vee \overline{v_{t,j}}\leftarrow\hspace{-10em}  & \text{for every }t\text{ in }T, r{\in}\Pi, x{\in}\at(\Pi),  0{\leq} i{<} \ceil{\log(\Card{\chi(t)})}, 1{\leq} j {\leq} 3 
	\\
%
%
%
%
%
	&\textbf{Saturate Pointers and Values}\hspace{-10em}\notag\\
	\label{redt:saturate}&b_{t,j}^i \leftarrow sat_{\rootOf(T)} \quad \overline{b_{t,j}^i} \leftarrow sat_{\rootOf(T)} \quad \leftarrow \neg sat_{\rootOf(T)}\hspace{-10em} &\text{for every }t\text{ in }T, 0{\,\leq\,} i{\,<\,} \ceil{\log(\Card{\chi(t)})}, 1{\,\leq\,}j{\,\leq\,}3\\
	%
	%
	%
	\label{redt:saturate2}&v_{t,j} \leftarrow sat_{\rootOf(T)}\quad  \overline{v_{t,j}} \leftarrow sat_{\rootOf(T)}\quad  sat_r \leftarrow sat_{\rootOf(T)} \quad \overline{sat_r}\leftarrow sat_{\rootOf(T)} \hspace{-10em} &\text{for every }t\text{ in }T, r\in\Pi, 1\leq j\leq 3\\
	%
	%
	%
	%
	%
%
%
	&\textbf{Synchronize Pointers and Values}\hspace{-10em}\notag\\
%
%
%
&sat_{\rootOf(T)} \leftarrow sat_r, \dot{b} 
&\hspace{-5em}\text{for every }t\text{ in }T, 
t=\varphi(r), x\in\at(r), j=\ord(r,x), 
b\in\bvali{x}{t}{j}
\label{redt:ptr}\\
\label{redt:satguess}&sat_{\rootOf(T)} \leftarrow b^0, \ldots, b^n, \overline{v_{t,j}}, x 
&\text{for every }t\text{ in }T, t=\varphi(x), 1\leq j\leq 3, \bvali{x}{t}{j}=\{b^0, \ldots, b^n\}\\
\label{redt:satguess2}&sat_{\rootOf(T)} \leftarrow b^0, \ldots, b^n, v_{t,j}, \overline{x} 
&\text{for every }t\text{ in }T, t=\varphi(x), 1\leq j\leq 3, \bvali{x}{t}{j}=\{b^0, \ldots, b^n\}\\
%
%
%
%
%
%
%
%
&sat_{\rootOf(T)} \leftarrow b_t^0, \ldots, b_t^n, \dot{b_{t'}}
&\text{for every }t\text{ in }T,  t'\in\children(t),x\in\chi(t)\cap\chi(t'), 1\leq j\leq 3, \notag\\
&sat_{\rootOf(T)} \leftarrow 
\overline{v_{t,j}}, v_{t',j} \qquad\qquad\quad sat_{\rootOf(T)} \leftarrow 
{v_{t,j}}, 
\overline{v_{t',j}}\hspace{-10em}
&\bvali{x}{t}{j}{=}\{b_t^0, \ldots, b_t^n\}, b_{t'}{\in}\bvali{x}{t'}{j} \label{redt:satval}\\
%
%
	&\textbf{Check Satisfiability of Rules}\hspace{-10em}\notag\\
&sat_{\rootOf(T)} \leftarrow sat_r, v_{t,j}  & \text{ for every }t\text{ in }T,  t=\varphi(r), 
x\in (H_r \cup B_r^-), 
	\label{redt:head-true} 
	j=\ord(r,x)\\
	%
	%
	&
	sat_{\rootOf(T)}\leftarrow 
	sat_r, \overline{v_{t,j}}  & \text{ for every }t\text{ in }T,  t=\varphi(r),x\in B_r^+, 
	\label{redt:bodyplus-false} 
	j=\ord(r,x)\\
	&sat_t \leftarrow sat_{t_1}, \ldots, sat_{t_o}, \overline{sat_{r_1}}, \ldots, \overline{sat_{r_m}}  & \hspace{-10em}\text{ for every }t\text{ in }T, \children(t){=}\{t_1,\ldots,t_o\}, \{r{\in} \Pi \mid t{=}\varphi(r)\}{=}\{r_1,\ldots,r_m\}  
	\label{redt:notchosen} 
\end{flalign}
}\caption{The reduction~$\mathcal{R}_{td}$ that takes a normalized fully tight program~$\Pi$ and an  annotated nice TD~$\mathcal{T}=(T,\chi,\varphi)$ of~$\mathcal{G}_{\Pi}$.}
\label{figt:redt}
\end{figure*}

\smallskip
\noindent\textbf{The Reduction.}
Conceptually, the ideas of our reduction~$\mathcal{R}_{td}$ for exponentially decreasing treewidth are similar to above. However, we require different auxiliary atoms.
To this end, consider a normalized fully tight program~$\Pi$ and an annotated nice TD~$\mathcal{T}{\,=\,}(T,\chi,\varphi)$ of~$\mathcal{G}_{\Pi}$.
Then, we use  bit atoms of the form~$b_{t,j}^i$ and~$\overline{b_{t,j}^i}$, i.e., we require three pointers for each node~$t$ of~$T$, where~$1{\,\leq\,} j{\,\leq\,} 3$ and~$0{\,\leq\,} i {\,<\,} \ceil{\log(\Card{\chi(t)})}$. A consistent subset over these bit atoms for a fixed~$t$ and~$j$ addresses (targets at) exactly one atom contained in~$\chi(t)$. Hence, to represent the value of this target, we require corresponding value atoms~$v_{t,j}$.

In order to avoid
unintentional increases of treewidth, we implicitly split Rule~(\ref{red:sats}) and guide it along~$\mathcal{T}$, from the leaves towards the root of~$T$. In addition to atoms~$sat_r, \overline{sat_r}$ for a rule~$r\in \Pi$, atoms~$sat_t$ store the evaluation of parts of Rule~(\ref{red:sats}) up to~$t$. So, we do not need atom~$sat$ and instead use~$sat_{t^*}$ where~$t^*{=}\rootOf(T)$.
Finally, for every atom~$x\in \chi(t)$, we let~$\bvali{x}{t}{j}$ be the set of bit atoms for the $j$-th pointer that uniquely addresses~$x$, assuming (as above) that~$\bvali{x}{t}{j}$ uniquely binary-encodes the ordinal position of~$x$ in~$\chi(t)$.

Reduction~$\mathcal{R}_{td}$ is given in Figure~\ref{figt:redt}. 
Compared to the previous reduction, we synchronize atoms~$x$ with pointer values in Rules~(\ref{redt:satguess}) and~(\ref{redt:satguess2}), where, assuming that $t{=}\varphi(x)$, we preserve the exponential decrease of treewidth by synchronizing only one atom in~$\chi(t)$ (i.e., not more than~$\log(k)$ bits) at once. 
Further, Rules~(\ref{redt:satval}) synchronize neighboring pointers (bit atoms) and bit value atoms. 
Satisfiability of a rule is verified per bag, 
using Rules~(\ref{redt:head-true}),	(\ref{redt:bodyplus-false}).
Finally, Rules~(\ref{redt:notchosen}) guide Rule~(\ref{red:sats}) along~$\mathcal{T}$, thereby evaluating whether all atoms $\overline{sat_r}$ hold for rules~$r\in\Pi$, where~$\varphi(r)$ maps to a node below~$t$.

\smallskip
\noindent\textbf{Results.} As above, we prove runtime and correctness~next.

\begin{lemma}[Runtime \& Correctness]\label{td:corr}
Let~$\Pi$ be any normalized fully tight program, $\mathcal{T}$ be an annotated nice TD of~$\mathcal{G}_{\Pi}$ of width~$k$, and let~$\Pi'=\mathcal{R}_{td}(\Pi, \mathcal{T})$. $\mathcal{R}_{td}$ runs in time $\mathcal{O}((\Card{\at(\Pi)}+\Card{\Pi})\cdot k)$ and is correct: Any answer set~$M$ of~$\Pi$ can be extended to an answer set~$M'$ of~$\Pi'$. Also, for any answer set~$M'$ of~$\Pi'$, $M'\cap\at(\Pi)$ is an answer set of~$\Pi$.
\end{lemma}

Indeed, $\mathcal{R}_{td}$ exponentially decreases treewidth.
\begin{lemma}[TW Compression]\label{tw:compr}
Let~$\Pi$ be any normalized fully tight program, $\mathcal{T}=(T,\chi,\varphi)$ be any annotated TD of~$\mathcal{G}_{\Pi}$ of width~$k$, and~$\Pi'{=}\mathcal{R}_{td}(\Pi,\mathcal{T})$. There is a TD of $\mathcal{G}_{\Pi''}$ of width $\mathcal{O}(\ceil{\log(k)})$, where~$\Pi''$ is a normalized program of $\Pi'$.
\end{lemma}

Since compression works for tree decompositions,
it also works for path decompositions, yielding the following result.

\begin{theorem}[PW/CLW LB]\label{thm:clw}
Let~$\Pi$ be any normalized disjunctive program. Under ETH the consistency of~$\Pi$ cannot be decided in time $2^{2^{o(k)}}\cdot\poly(\Card{\at(\Pi)})$, where~$k$ is the pathwidth or the cliquewidth of~$\mathcal{G}_{\Pi}$ or $\mathcal{I}_\Pi$.
\end{theorem}
%

Further, we can show stronger bounds than pathwidth by linking a more restricted variant of pathwidth to bandwidth.

\begin{lemma}\label{lem:pwbw}
Let~$G{\,=\,}(V,E)$ be a graph, $\mathcal{T}{\,=\,}(T,\chi)$ be a PD of $G$ of width~$k$,
s.t.\ every vertex in~$V$ occurs in at most two  bags of~$\mathcal{T}$.
Then, the bandwidth of~$G$ is bounded by~$2k-1$.
\end{lemma}
\begin{theorem}[BW/CW LB]\label{thm:bw}
Let~$\Pi$ be any normalized~disjunctive program. Under ETH, consistency of~$\Pi$ cannot be~decided in $2^{2^{o(k)}}\cdot\poly(\Card{\at(\Pi)})$ for bandwidth~$k$ of~$\mathcal{G}_\Pi$~($\mathcal{I}_\Pi$).
\end{theorem}

While bandwidth bounds the degree 
to obtain a tight bound for cutwidth, we require a \emph{constant} degree (independent from cutwidth).
%
%
%
However, we 
add auxiliary atom copies to ensure constant degree while  \emph{linearly} preserving the parameter.

\begin{theorem}[CW LB]\label{thm:cwlb}
Let~$\Pi$ be any normalized disjunctive program. Under ETH the consistency of~$\Pi$ cannot be decided in time $2^{2^{o(k)}}\cdot\poly(\Card{\at(\Pi)})$ for cutwidth~$k$ of~$\mathcal{G}_\Pi$ or $\mathcal{I}_\Pi$.
\end{theorem}

\section*{Discussion and Conclusion}
In this paper we discussed the computational effort
of disjunctive \ASP in terms of structural measures.
Unfortunately, for 
\emph{treewidth}, under reasonable assumptions (ETH), there is no 
hope 
for a runtime significantly better
than \emph{double-exponential} in treewidth.
What about {other} 
parameters?

It turns out that for normalized programs and 
if we take a vertex cover of the program's rule structure ({incidence graph}), 
we obtain a kernel that is polynomial in the vertex cover's size.
Note that this even holds despite the implicit subset-minimization,
which, when translating into a logical formula causes the duplication of
atoms, i.e., the formula admits only huge vertex covers.
Hence, we achieve a \emph{single-exponential} algorithm for disjunctive \ASP.
While this is good news, some practical programs admit only huge vertex covers in
their structure. 
Is there a useful measure smaller than vertex cover that still admits a single-exponential algorithm?

We develop a new technique that allows us to reduce from non-disjunctive \ASP to disjunctive \ASP, thereby exponentially reducing structural dependencies. With this approach we trade a significantly simpler structure for additional computational hardness. Unfortunately, under ETH,  we 
give negative answers by excluding most prominent structural parameters between vertex cover and treewidth: treedepth, feedback vertex size, cliquewidth, pathwidth, bandwidth, and cutwidth. 

These results yield insights into the hardness of disjunction, thereby providing simple program structures that are already ``hard''.
Indeed, from a practical point of view, this renders disjunction significantly harder,
already for simple structural dependencies.
In fact, disjunction allows us to simplify structure.
%
So, how much does structure really help? We 
\emph{hardly} expect 
better results for 
measures smaller than a vertex~cover. 

\smallskip
\noindent \textbf{Future Work. }This paper opens up several questions for future research.
Combining structural and non-structural measures may be
strong enough to better confine disjunction. 
%
%
This paper provides a starting point towards a
modeling guideline for limiting the cost of disjunction. We expect progress in this direction in the future.
Further, seeing how our bounds persist for 
stronger versions of ETH, e.g., SETH, is interesting.
%
Do our reductions pay off in practice? Since the ones in Figures~\ref{fig:red} and~\ref{figt:redt} preserve answer sets, we expect use cases for counting-like problems. 
For neuro-symbolic applications, where, (algebraic) counting is the basis for parameter learning, we expect 
improvements by 
our ideas. 
Here, 
as demonstrated 
by 
counting competitions~\cite{FichteHecherHamiti20,KorhonenMatti21},
using structure is crucial.


\clearpage
\section*{Acknowledgments}
Authors are ordered alphabetically.
Part of the work has been carried out while Hecher visited the
Simons Institute at UC Berkeley.
Research is supported by 
%
%
the Austrian Science Fund (FWF), grants J4656 and P32830, and the Society for
Research Funding in Lower Austria (GFF) grant ExzF-0004.

\bibliography{references}

\clearpage
\appendix

\section*{Additional Proofs of Section~\ref{sec:vc}}

\begin{restatetheorem}[thm:primal_vc_kernel]
\begin{theorem}[Polynomial VC-Kernel]
Let $\Pi$ be a program such that for every rule $r \in \Pi$ we have $|H_r| + |B_r^-| + |B_r^+| \leq c$ for~$c \in \mathbb{N}$ and let $S \subseteq \at(\Pi)$ be a vertex cover of $\mathcal{G}_{\Pi}$.
Then there is a program $\Pi'$ s.t.\ (i) $|\at(\Pi')| \leq 4^{c}\binom{|S|}{c}$ and (ii) $\Pi$ is consistent iff $\Pi'$ is consistent.
\end{theorem}
\end{restatetheorem}
For the proof, we use the \emph{roles} $R(a, Pos, \Pi)$, for $Pos = H, B^+, B^-$, of atoms $a \in \at(\Pi) \setminus S$, where $R(a, Pos, \Pi)$ is the set of rules $r'$ such that $H_{r'} = H_r \cap S, B_{r'}^+ = B_r^+ \cap S, B_{r'}^- = B_r^- \cap S$ for some rule $r \in \Pi$ with $a \in Pos_r$. I.e.\ the set of rules that can be obtained from a rule in $\Pi$ that uses $a$ by removing $a$.
\begin{proof}
Let $\Pi$, $c$, and $S$ as in the statement of the theorem.

For two atoms $a, b \in \at(\Pi)$, we write $a \sim b$ iff $R(a, Pos, \Pi) = R(b, Pos, \Pi)$ for $Pos = H, B^+, B^-$.

Observe, that the number of atoms $a, b \in \at(\Pi)$ such that $a \not\sim b$ is bounded by a term that only depends on $|S|$ and $c$. We can see this as follows: The rules in $R(.,.,\Pi)$ only use atoms from $S$, since otherwise $S$ is not a vertex cover. Every rule $r$ in $\Pi$ that uses $a$ has at most $c-1$ other atoms. There are $\binom{|S|}{c-1}$ subsets $S'$ of $S$ such that $|S'| = c - 1$. Using only atoms from $S' \cup \{a\}$ we can construct at most $4^{c}$ different rules, since w.l.o.g.\ each of the $c$ atoms can independently occur only in exactly one of four ways, namely, (i) in the head, (ii) in the posivitive body, (iii) in the negative body, and (iv) none of the above.\footnote{If an atom occurs more than once we can remove the rule of rewrite it such that it does not occur.} 
Together, this results in a bound of $4^c\binom{|S|}{c}$ on the number of atoms that can be distinguished based on $R(a, Pos, \Pi)$ alone.

The kernelization is based on the fact that if for two atoms $a, b \in \at(\Pi)$ we have $a \sim b$, then we can remove one of them from $\Pi$ without changing $\Pi$'s consistency. Intuitively, this is easy to understand: $R(a,.,\Pi)$ determines the role that $a$ has in $\Pi$, so if there are two atoms that have the same role, then one of them is redundant.

Formally, we proceed in two steps. First, we prove that if $a \sim b$ and $\mathcal{I} \subseteq \at(\Pi)$ then (i) if $a \in \mathcal{I}$, then $\mathcal{I} \cup \{b\} \models \Pi$ iff $\mathcal{I} \models \Pi$, and (ii) if $a \not\in \mathcal{I}$, then  $\mathcal{I} \setminus \{b\} \models \Pi$ iff $\mathcal{I} \models \Pi$. 

For (i) assume that on the contrary $\mathcal{I} \cup \{b\} \not\models \Pi$ and $\mathcal{I} \models \Pi$. Then there exists a rule $r \in \Pi$ such that $(H_r \cup B_r^-) \cap (\mathcal{I} \cup \{b\}) = \emptyset$ and $B_r^+ \setminus (\mathcal{I} \cup \{b\}) = \emptyset$. Since $\mathcal{I}$ satisfies $r$ and $\mathcal{I} \subseteq \mathcal{I} \cup \{b\}$, we know that $B_r^+ \setminus \mathcal{I} \neq \emptyset$. More precisely, $B_r^+ \setminus \mathcal{I} = \{b\}$ follows. However, since $a \sim b$, we know that there exists a rule $r'$ such that $B_{r'}^+ = B_{r} \setminus \{b\} \cup \{a\}$ and $H_r = H_{r'}, B_r^- = B_{r'}^-$. This leads to a contradiction, since $a \in \mathcal{I}$ but $\mathcal{I}$ satisfies $r'$.

On the other hand, assume that $\mathcal{I} \cup \{b\} \models \Pi$ and $\mathcal{I} \not\models \Pi$. Then there exists a rule $r \in \Pi$ such that $(H_r \cup B_r^-) \cap \mathcal{I} = \emptyset$ and $B_r^+ \setminus \mathcal{I} = \emptyset$. Since $\mathcal{I} \cup \{b\}$ satisfies $r$ and $\mathcal{I} \subseteq \mathcal{I} \cup \{b\}$, we know that $(H_r \cup B_r^-) \cap (\mathcal{I} \cup \{b\}) \neq \emptyset$. More precisely, $(H_r \cup B_r^-) \cap (\mathcal{I} \cup \{b\}) = \{b\}$ follows. However, since $a \sim b$, we know that there exists a rule $r'$ such that $(H_{r'} \cup B_{r'}^-) = (H_r \cup B_r^-) \setminus \{b\} \cup \{a\}$ and $B_r^+ = B_{r'}^+$. This leads to a contradiction, since $a \in \mathcal{I}$ but $\mathcal{I}$ does not satisfy $r'$.

We can prove (ii) in a similar manner.

Second, we prove that if $a \sim b$, $\mathcal{I} \subseteq \at(\Pi)$, then (i) if $a \in \mathcal{I}$, then $\mathcal{I} \setminus \{b\}$ is not an answer set of $\Pi$ and (ii), by symmetry of $\sim$, if $b \in \mathcal{I}$, then $\mathcal{I} \setminus \{a\}$ is not an answer set of $\Pi$. We know that if $\mathcal{I}$ is an answer set of $\Pi$, then there does not exist an interpretation $\mathcal{I}' \subsetneq \mathcal{I}$ such that $\mathcal{I}' \models \Pi$. However, if 
$\mathcal{I} \setminus \{b\}$ is an answer set of $\Pi$, then by the first step, $\mathcal{I} \setminus \{a,b\}$ satisfies $\Pi$. This is a contradiction to the subset minimality of answer sets. 

This shows that we are not loosing answer sets by removing all rules in which $b$ occurs from $\Pi$ obtaining $\Pi'$. Clearly, we also do not gain new answer sets, since any answer set $\mathcal{I}$ of $\Pi'$ either (i) $\mathcal{I}$ is already an answer set of $\Pi$ if $a \not\in \mathcal{I}$ or (ii) $\mathcal{I} \cup \{b\}$ is an answer set of $\Pi$.
\end{proof}

\begin{restatecorollary}[cor:vcext]
\begin{corollary}
Let $\Pi$ be a program such that for every rule $r \in \Pi$ we have $|H_r| + |B_r^-| + |B_r^+| \leq c$ for~$c \in \mathbb{N}$, where every atom has at most one neighbor in $\mathcal{G}_{\Pi}^1$ or $\mathcal{G}_{\Pi}^2$ (that is not a neighbor in $\mathcal{G}_{Pi}^0$). Furthermore, let $S \subseteq \at(\Pi)$ be a vertex cover of $\mathcal{I}_{\Pi}^0$.
Then there exists a program $\Pi'$ such that (i) $|\at(\Pi')| \leq 3 \cdot 4 \cdot (4^{c}\binom{|S|}{c})^2 + 4^{c}\binom{|S|}{c}$ and (ii) $\Pi$ is consistent iff $\Pi'$ is consistent.
\end{corollary}
\end{restatecorollary}
\begin{proof}
Here, we need to redefine the equivalence relation $\sim$ between atoms $a \in \at(\Pi) \setminus S$ as follows. Namely, $a \sim b$ iff (i) $a$ occurs in a type 1 rule iff $b$ occurs in a type 1 rule, (ii) $a$ occurs in a type 2 rule iff $b$ occurs in a type 2 rule, and (iii), the have the same \emph{compound role}. Here, the compound role of atoms $a$ that occur in a type 1 or 2 rule with second atom $a' \not\in S$ the compound role is the combination of the roles of $a$ and $a'$ and whether there is only one rule such that $a$ and $a'$ cooccur or two. Otherwise, the compound role of $a$ is the role of $a$.

For atoms that do not occur in a type 1 or 2 rule or whose second atom $a'$ is in $S$, we can proceed as in Theorem \ref{thm:primal_vc_kernel}. This leads to at most $4^{c}\binom{|S|}{c}$ atoms.

The remaining atoms $a \not\in S$ occur either in a type 1 or 2 rule with second atom $a'$. Due to analogous reasoning as in Theorem \ref{thm:primal_vc_kernel}, there are at most $(4^{c}\binom{|S|}{c})^2$ atoms $a$ that occur in a type 1 (resp.\ type 2) rule that are not equivalent w.r.t.\ $\sim$. For atoms $a_1, \dots, a_n$ that are equivalent w.r.t.\ $\sim$, we can remove all rules that contain $a_5, ...,$ or $a_n$. By keeping four representatives $a_1, \dots, a_4$ (with respective second atoms $a_1', \dots, a_4'$, we keep all possible combinations of truth values for $a, a'$ that could occur together open. This results in at most 2 plus 1 (for type 1 or type 2, respectively) times $4$ (for each combination of truth values) times $(4^{c}\binom{|S|}{c})^2$ (for each different compound role) atoms that we cannot remove from the program.
\end{proof}

\begin{restatecorollary}[cor:vcinc]
\begin{corollary}
Let $\Pi$ be a program such that for every rule $r \in \Pi$ we have $|H_r| + |B_r^-| + |B_r^+| \leq c$ for~$c \in \mathbb{N}$ and let $S \subseteq \at(\Pi)$ be a vertex cover of $\mathcal{I}_{\Pi}$.
Then there exists a program $\Pi'$ such that (i) $|\at(\Pi')| \leq 4^{c}\binom{|S|}{c}c^c$ and (ii) $\Pi$ is consistent iff $\Pi'$ is consistent. 
\end{corollary}
\end{restatecorollary}
\begin{proof}
We can replace every rule $r \in \Pi$, by the two rules $H_r \leftarrow aux_r$ and $aux_r \leftarrow B_r^+, B_r^-$, where $aux_r$ is a fresh atom, obtaining a new program $\Pi_I$. We can obtain a vertex cover $S'$ for $\mathcal{G}_{\Pi_I}$ as 
\(
S' = S \cup \at(\Pi) \cup \bigcup_{r \in S} H_r \cup B_r^+ \cup B_r^-.
\)
Then $|S'| \leq |S|\cdot c$, since for each rule $r$ in the original cover we add at most $c$ atoms to $S'$. Finally, we obtain the desired result by applying Theorem \ref{thm:primal_vc_kernel} to $\Pi_I$ with vertex cover $S'$.
\end{proof}

\begin{restatecorollary}[cor:vcruntime]
\begin{corollary}
Let $\Pi$ be a program such that for every rule $r \in \Pi$ we have $|H_r| + |B_r^-| + |B_r^+| \leq c$ for~$c \in \mathbb{N}$ and vertex cover size $k$ of the primal graph.
Then we can decide consistency of $\Pi$ in time $\mathcal{O}(2^{2\cdot4^{c}\binom{k}{c}}\poly(|\at(\Pi)|))$.
\end{corollary}
\begin{proof}
First, we obtain a vertex cover $S$ of size $k$ for the primal graph using the algorithm from \cite{niedermeier1999upper}, which runs in time $\mathcal{O}(k\Card{\at(\Pi)} + (1,29175)^k \cdot k^2)$. Then we apply our kernelization result in polynomial time in $\Pi$, by computing the roles of all atoms and removing atoms with duplicate roles. Finally, since the number of atoms is bounded by $4^c\binom{k}{c}$, we can check consistency by iterating over all $2^{4^c\binom{k}{c}}$ interpretations $\mathcal{I}$. If $\mathcal{I}$ satisfies $\Pi$, we further iterate over at most $2^{4^c\binom{k}{c}}$ interpretations $\mathcal{I}' \subsetneq \mathcal{I}$ to check if $\mathcal{I}$ is a $\subseteq$-minimal satisfying interpretation of $\Pi^{\mathcal{I}}$.
\end{proof}
\end{restatecorollary}

\section*{Additional Proofs of Section~\ref{sec:fvs}}

\begin{restatelemma}[lem:rtmcorr]
\begin{lemma}[Runtime \& Correctness]
Let~$\Pi$ be any normalized fully tight program, $S$ be any FVS of~$\mathcal{G}_{\Pi}$, and let~$\Pi'{=}\mathcal{R}_{fvs}(\Pi, S)$. $\mathcal{R}_{fvs}$ runs in time~$\mathcal{O}(\Card{\at(\Pi)}+\Card{\Pi}\cdot\log(\Card{S})+\Card{S})$ and it is correct: Any answer set~$M$ of~$\Pi$ can be extended to an answer set~$M'$ of~$\Pi'$. Further, for any answer set~$M'$ of~$\Pi'$, $M'\cap\at(\Pi)$ is an answer set of~$\Pi$.
\end{lemma}
\end{restatelemma}
\begin{proof}
The runtime follows by careful analysis of Figure~\ref{fig:red}.

For correctness, we sketch both directions. ``$\Longrightarrow$'': 
Let~$M$ be an answer set of~$\Pi$. From this we construct~$M'\eqdef M \cup \{\overline{a}\mid a\in \at(\Pi)\setminus M\} \cup \{sat, sat_r, \overline{sat_r} \mid r\in \Pi\} \cup \{b_j^i, \overline{b_j^i}, v_j, \overline{v_j} \mid 0\leq i < \ceil{\log(\Card{S})}, 1\leq j \leq 3\}$.
Assume towards a contradiction that~$M'$ is not an answer set of~$\Pi'$.
Obviously, $M'$ is a model of~$\Pi'$, i.e., there has to exist a model~$M''\subsetneq M'$ of~${\Pi'}^{M'}$.

This can only occur if~$sat \notin M''$, since otherwise due to Rules~(\ref{red:saturate}), (\ref{red:keepsat}), $M''$ would be identical to~$M'$.
Then, there can not exist a rule~$r\in \Pi$ with~$sat_r\in M''$,
since~$M$ is by definition a model of~$\{r\}$.
Indeed, Rules~(\ref{red:bodyneg-false1}), (\ref{red:bodyneg-false2})
would then ensure that~$sat\in M''$ for atoms in~$r$ not in~$S$.
For those atoms in~$r$ that are also in~$S$,
Rules~(\ref{red:head-true}), (\ref{red:bodyplus-false})
ensure that~$sat\in M''$, since Rules~(\ref{red:satptr}) and (\ref{red:satguess}) take care of the equivalence between those atoms and atoms~$v_j$.

Consequently, there does not exist a rule~$r\in \Pi$ with~$sat_r\in M''$.
However, by Rule~(\ref{red:sats}), we conclude~$sat\in M''$, which contradicts that~$M''$ exists.

 ``$\Longleftarrow$'': Assume~$M'$ is an answer set of~$\Pi'$. We define~$M\eqdef M'\cap \at(\Pi)$. Suppose towards a contradiction that~$M$ is not an answer set of~$\Pi$.
 %
 %
 %
 Then, since~$\Pi$ is fully tight, there is a rule~$r\in\Pi$ such that~$M\not\models \{r\}$. 
 From this, we will construct a model~$M''\subsetneq M'$
 that is a model of~$\Pi^{M'}$, which will contradict that~$M'$ is an answer set of~$\Pi'$.
 More precisely, let~$M''\eqdef M \cup \{sat_r, \overline{sat_{r'}} \mid r'\in \Pi, r'\neq r\} \cup (\bigcup_{x\in\at(r)} \bval{x}{\ord(r,x)} \cup \{v_{\ord(r,x)} \mid x\in M \cap \at(r)\}$ $\cup \{\overline{v_{\ord(r,x)}}, \overline{x} \mid x\in \at(r)\setminus M\} \cup \{ \overline{b_j^i}, \overline{v_j} \mid \Card{\at(r)}< j\leq 3, 0\leq i <\ceil{\log(\Card{S})}\}$.
 By construction, $M''$ satisfies Rules~(\ref{red:guess}).
 Since~$sat\not\in M''$, $\leftarrow \neg sat$ is not in $\Pi^{M'}$,  $M''$ satisfies those Rules~(\ref{red:saturate}) and~(\ref{red:keepsat}) that are in~$\Pi^{M'}$. 
 Further, $M''$ also ensures that Rules~(\ref{red:satptr}), (\ref{red:satguess}), and (\ref{red:sats}) are satisfied.
 Finally, since~$M\not\models\{r\}$,
 $M''$ satisfies Rules~(\ref{red:head-true})--(\ref{red:bodyneg-false2}).
 Consequently, $M'$ is not an answer set of~$\Pi'$.
\end{proof}

\begin{restatetheorem}[thm:compr]
\begin{theorem}
Let~$\Pi$ be any normalized fully tight program, $S$ be any sparse FVS of~$\mathcal{G}_{\Pi}$, and let~$\Pi'=\mathcal{R}_{fvs}(\Pi, S)$. Then, there is a sparse FVS of
$\mathcal{G}_{\Pi''}$ of size~$\mathcal{O}(\ceil{\log(\Card{S})})$, where~$\Pi''$ is a normalized program of~$\Pi'\setminus\{r\}$, with $r$ being Rule~(\ref{red:sats}).
\end{theorem}
\end{restatetheorem}

\begin{proof}
Let~$S{'\eqdef\,} \{sat, v_j, \overline{v_j}, b^i_j, \overline{b^i_j}\mid 0\leq i <\ceil{\log(\Card{S})},\allowbreak 1\leq j\leq 3\}$, which is of size~$6\ceil{\log(\Card{S})}+7$ 
It will be the feedback vertex set of~$\mathcal{G}_{\Pi''}$,
where $\Pi''$ is a normalized program of~$\Pi'\setminus\{r\}$.
Observe that indeed every cycle in~$\mathcal{G}_{\Pi'}$ due to Rules~(\ref{red:guess})--(\ref{red:bodyplus-false}) contains at least on atom in~$S'$.
Further, since~$S$ is sparse, for every two different atoms~$x,y\in\at(\Pi)$, there
is at most one unique rule~$r'\in\Pi$ with~$x,y\in \at(r')$.
As a result, Rules~(\ref{red:bodyneg-false1}) and~(\ref{red:bodyneg-false2}) do not cause a cycle in~$\mathcal{G}_{\Pi}$ that does not contain~$sat$, which is contained in~$S'$.
It remains to establish how the normalized program~$\Pi''$ is obtained, such that~$S'$ is a FVS of~$\mathcal{G}_{\Pi''}$.
The only non-normalized rules in~$\Pi'\setminus\{r\}$ are~(\ref{red:satguess}).
The known standard reduction~\cite{Truszczynski11} suffices, where for every body atom~$b$ creates an auxiliary rule with a fresh auxiliary atom in its head, as well as~$b$ and the previously introduced atom in its body.
For Rules (\ref{red:satguess}) and~$x{\,\in\,} S$, auxiliary rules for $x$ and $\overline{x}$ are constructed first, avoiding cycles in~$\mathcal{G}_{\Pi''}$ over~$3$ atoms not in~$S'$.
\end{proof}

\begin{restatetheorem}[thm:almostpath]
\begin{theorem}
Let~$\Pi$ be any disjunctive program. Under ETH, the consistency of~$\Pi$ cannot be decided in time $2^{2^{o(k)}}\cdot\poly(\Card{\at(\Pi)})$ for distance~$k$ to almost paths of $\mathcal{I}_\Pi$. The result still holds if the largest path length is in $\mathcal{O}(k)$.
\end{theorem}
\end{restatetheorem}
\begin{proof}
Let~$\Pi'$ be a normalized fully tight program. If we use~$\mathcal{R}_{fvs}$ on~$\Pi'$ and the set~$S=\at(\Pi')$, we obtain~$\Pi$. Let~$G$ be the graph obtained from $\mathcal{I}_\Pi$ after removing~$S$. 
Assume we obtain~$S''$ as in Lemma~\ref{lem:compr:fs}.
For every~$s\in S''$, graph~$G$ contains a path~$s,r,\overline{s}$, where~$r$ is due to Rules~(\ref{red:guess}). Further, due to Rules~(\ref{red:satguess}), both~$s$ and~$\overline{s}$ have an additional vertex of degree~$1$ as a neighbor.
Similarly, for vertices of the form~$sat_{r'}$, $G$ contains a path of the form~$sat_{r'},r'',\overline{sat_{r'}}$ due to Rules~(\ref{red:guess}). Then, due to Rules~(\ref{red:satptr}), (\ref{red:head-true}), and (\ref{red:bodyplus-false}), vertex $sat_{r'}$ has additional, (up to~$3k+6$ many) degree-$1$ vertices as neighbors.
In both cases, we can modify~$\Pi$ and construct~$\Pi''$, where we use fresh auxiliary copies 
${sat_{r'}}_1, \ldots$ of~$sat_{r'}$, 
such that each of these copies appears in exactly one of the Rules~(\ref{red:satptr}), 
(\ref{red:head-true}), and 
(\ref{red:bodyplus-false}). Note that thereby we also add to~$\Pi''$ rules of the form 
${sat_{r'}}_{i+1} \leftarrow {sat_{r'}}_i$ (as well as 
${sat_{r'}}_1\leftarrow sat_{r'}$) are required. 
Consequently, instead of~$\Pi$, we obtain a program~$\Pi''$ such that the graph obtained from $\mathcal{I}_{\Pi''}$ by removing $S''$, consists of paths of length~$\mathcal{O}(k)$, where every element of the path can have up to one additional neighbor vertex of degree~$1$.
\end{proof}

\begin{restatetheorem}[lem:compr:tdp]
\begin{theorem}[TDP LB]
Let~$\Pi$ be any disjunctive program. Under ETH the consistency of~$\Pi$ cannot be decided in time $2^{2^{o(k)}}\cdot\poly(\Card{\at(\Pi)})$ for treedepth~$k$ of $\mathcal{I}_\Pi$.
\end{theorem}
\end{restatetheorem}
\begin{proof}
%
%
%
Assume towards a contradiction that despite ETH, we can decide~$\Pi$ in time $2^{2^{o(k)}}\cdot\poly(\Card{\at(\Pi)})$. 
As in Theorem~\ref{thm:lb:fvs}, we take any 3-CNF formula~$F$ and transform it into a  normalized fully tight 
program~$\Pi'$.
%
%
%
%
Then, we apply our reduction and construct~$\Pi\eqdef\mathcal{R}_{fvs}(\Pi', \at(\Pi'))$, running in polynomial time. 
By Lemma~\ref{lem:compr:fs}, there is a FVS~$S'$ of~$\mathcal{I}_{\Pi}$ of size~$\mathcal{O}(k)$ with~$k\eqdef\log(\Card{\at(\Pi')})$. Even further, in the construction of~$S'$ in Theorem~\ref{thm:compr}, the longest path in the graph obtained from~$\mathcal{I}_{\Pi}$ after removing~$S'$, is of (constant) length~$4$. Two edges of such a path in~$\mathcal{I}_{\Pi}$ are due to Rule~(\ref{red:guess}); two extra edges are due to rule vertices by Rules~(\ref{red:satguess}). 
Consequently, the treedepth of~$\mathcal{I}_{\Pi}$ is obviously bounded by~$k$.
Solving~$\Pi$ in time~$2^{2^{o(k)}}\cdot\poly(\Card{\at(\Pi')})$ implies solving~$F$ in time~$2^{o(\Card{\var(F)})}\cdot\poly(\Card{\var(F)})$, contradicting ETH.
\end{proof}

\section*{Additional Proofs of Section~\ref{sec:bw}}

\begin{restatelemma}[td:corr]
\begin{lemma}[Runtime \& Correctness]
Let~$\Pi$ be any normalized fully tight program, $\mathcal{T}=(T,\chi, \varphi)$ be an annotated nice TD of~$\mathcal{G}_{\Pi}$ of width~$k$, and let~$\Pi'=\mathcal{R}_{td}(\Pi, \mathcal{T})$. $\mathcal{R}_{td}$ runs in time $\mathcal{O}((\Card{\at(\Pi)}+\Card{\Pi})\cdot{k})$ and is correct: Any answer set~$M$ of~$\Pi$ can be extended to an answer set~$M'$ of~$\Pi'$. Also, for any answer set~$M'$ of~$\Pi'$, $M'\cap\at(\Pi)$ is an answer set of~$\Pi$.
\end{lemma}
\end{restatelemma}
\begin{proof}
The runtime results follow from careful analysis of Figure~\ref{figt:redt} and the observation that an annotated nice TD requires
$\mathcal{O}(\Card{\at(\Pi)}+\Card{\Pi})$ many nodes, i.e., redundant nodes can be removed.

For correctness, we sketch both directions. ``$\Longrightarrow$'': 
Let~$M$ be an answer set of~$\Pi$. 
Similar to Lemma~\ref{lem:rtmcorr}, we construct~$M'\eqdef M \cup \{\overline{a}\mid a\in \at(\Pi)\setminus M\}\allowbreak \cup \{sat_r, \overline{sat_r}, sat_t, \overline{sat_t}, b_{t,j}^i, \overline{b_{t,j}^i}, v_{t,j}, \overline{v_{t,j}} \mid t\text{ in }T, r\in\Pi, 0\leq i < \ceil{\log(\Card{S})}, 1\leq j \leq 3\}$.
Assume towards a contradiction that~$M'$ is not an answer set of~$\Pi'$.
Obviously, $M'$ is a model of~$\Pi'$, i.e., there has to exist a model~$M''\subsetneq M'$ of~${\Pi'}^{M'}$.
This can only occur if~$sat_{\rootOf(T)} \notin M''$, since otherwise due to Rules~(\ref{redt:saturate}), (\ref{redt:saturate2}), $M''$ would be identical to~$M'$. 
%
%
By Rules~(\ref{redt:notchosen}), we know that
there has to exist at least one rule~$r\in \Pi$
such that~$sat_r\in M''$.
Indeed, if this was not the case, $sat_{\rootOf(T)}\in M''$. 
%
%
Then, by Rules~(\ref{redt:ptr}) 
there is a node~$t=\varphi(r)$ with~$\bvali{a_j}{t}{j}\subseteq M''$ for every~$1\leq j\leq \Card{\at(r)}$ such that~$a_j\in\var(r)$ and~$j=\ord(r,a_j)$. 
%
%
By Rules~(\ref{redt:satval}) for every node~$t'$ with~$a_j\in \chi(t')$ such that~$1\leq j\leq \Card{\at(r)}$, we have~$\bvali{a_j}{t'}{j}\subseteq M''$.
As a result, by Rules~(\ref{redt:satguess}) and~(\ref{redt:satguess2}), $a_j\in M''$ iff $v_{t,j}\in M''$.
Consequently, due to Rules~(\ref{redt:head-true}) and~(\ref{redt:bodyplus-false}) and since $M$ is a model of~$\Pi$, we conclude that~$sat_{\rootOf(T)}\in M''$.
Consequently, there does not exist a rule~$r\in \Pi$ with~$sat_r\in M''$, which contradicts that~$M''$ exists.

``$\Longleftarrow$'': Suppose~$M'$ is an answer set of~$\Pi'$. We define~$M\eqdef M'\cap \at(\Pi)$. Assume towards a contradiction that~$M$ is not an answer set of~$\Pi$.
Then, since~$\Pi$ is fully tight, there is at least one rule~$r\in \Pi$ such that $M\not\models \{r\}$.
From this, we will construct a model~$M''\subsetneq M'$
 that is a model of~$\Pi^{M'}$, which will contradict that~$M'$ is an answer set of~$\Pi'$.
 More precisely, let~$M''\eqdef M \cup \{sat_r, \overline{sat_{r'}} \mid r'\in \Pi, r'\neq r\} \cup \{b\mid t\text{ in }T, x\in\at(r)\cap\chi(t), b\in\bvali{x}{t}{\ord(r,x)}\} \cup \{v_{t,\ord(r,x)} \mid x\in M \cap \at(r), t\text{ in }T\}$ $\cup \{\overline{v_{t,\ord(r,x)}}, \overline{x} \mid x\in \at(r)\setminus M, t\text{ in }T\} \cup \{ \overline{b_{t,j}^i}, \overline{v_{t,j}} \mid \Card{\at(r)}< j\leq 3, 0\leq i <\ceil{\log(\Card{\chi(t)})}, t\text{ in }T\} \cup \{sat_t \mid t\text{ in }T, M\models\{r'\in \Pi \mid \varphi(r) \text{ is }t\text{ or a node below }t \text{ in } T\}\}$.
 By construction of~$M''$, Rules~(\ref{redt:guess}) are satisfied and Rules~(\ref{redt:saturate}),(\ref{redt:saturate2}) that are contained in~${\Pi}^{M'}$ are vacuously satisfied.
Further, also Rules~
(\ref{redt:ptr})--(\ref{redt:satval}) as well as Rules~(\ref{redt:notchosen})
are satisfied by~$M''$. 
Then, since  $M\not\models \{r\}$, $M''$ is also a model of Rules~(\ref{redt:head-true}) and~(\ref{redt:bodyplus-false}).
As a result, $M'$ is not an answer set of~$\Pi'$.
\end{proof}

\begin{restatelemma}[tw:compr]
\begin{lemma}[TW Compression]
Let~$\Pi$ be any normalized fully tight program, $\mathcal{T}=(T,\chi,\varphi)$ be any annotated TD of~$\mathcal{G}_{\Pi}$ of width~$k$, and~$\Pi'{=}\mathcal{R}_{td}(\Pi,\mathcal{T})$. There is a TD of $\mathcal{G}_{\Pi''}$ of width $\mathcal{O}(\ceil{\log(k)})$, where~$\Pi''$ is a normalized program of $\Pi'$.
\end{lemma}
\end{restatelemma}
\begin{proof}
We construct a TD~$\mathcal{T}'{\eqdef}(T,\chi')$ of~$\mathcal{G}_{\Pi'}$ by defining $\chi'$. For every~$t$ in~$T$, we let~$\chi'(t){\eqdef}\{sat_{\rootOf(T)}, sat_t,\allowbreak b_{t,j}^i, \overline{b_{t,j}^i} v_{t,j}, \overline{v_{t,j}}, sat_{t'}, b_{t',j}^{i'}, \overline{b_{t',j}^{i'}}, v_{t',j}, \overline{v_{t',j}} \mid t'\in\children(t),\allowbreak 1\leq j\leq 3, 0\leq i < \ceil{\log(\Card{\chi(t)})}, 0\leq i' < \ceil{\log(\Card{\chi(t')})}\} \cup \{sat_r, \overline{sat_r} \mid r\in \Pi, t=\varphi(r)\} \cup \{a, \overline{a} \mid a \in\chi(t), t=\varphi(a)\}$.
Since~$\Card{\children(t)}\leq 2$, we have~$\Card{\chi'(t)} \leq 3\cdot 6\cdot \ceil{\log(k)}+12$.
Further, indeed the atoms of every rule constructed by~$\mathcal{R}_{td}$, see Figure~\ref{figt:redt}, appears in at least one bag~$\chi'(t)$. Also, by construction, $\mathcal{T}'$ is connected, i.e., $\mathcal{T}'$ is indeed a valid TD.
We obtain~$\Pi''$ using the standard reduction~\cite{Truszczynski11} on~$\Pi'$, similar to Lemma~\ref{thm:compr}. Observe that the rules requiring normalization are~(\ref{redt:satguess})--(\ref{redt:satval}) and~(\ref{redt:notchosen}).
Also, the atoms of a such a rule appear in at least one bag~$\chi'(t)$ for a node~$t$ of~$T$.
Therefore, we obtain $\mathcal{T}''$ by modifying~$\mathcal{T}'$ (nodes may require duplication), such that every bag in~$\mathcal{T}''$ gets additional auxiliary atoms
due to normalization of \emph{at most one rule} in~$\Pi''$.
Hence, the width of~$\mathcal{T}''$ is~$\leq 20\ceil{\log(k)}+14$,
as the increase by~$2\ceil{\log(k)}+2$ is due to~(\ref{redt:satval}).
\end{proof}

\begin{restatetheorem}[thm:clw]
\begin{theorem}[PW/CLW LB]
Let~$\Pi$ be any normalized disjunctive program. Under ETH the consistency of~$\Pi$ cannot be decided in time $2^{2^{o(k)}}\cdot\poly(\Card{\at(\Pi)})$, where~$k$ is the pathwidth or the cliquewidth of~$\mathcal{G}_{\Pi}$ or $\mathcal{I}_\Pi$.
\end{theorem}
\end{restatetheorem}
\begin{proof}
Since every path decomposition is a tree decomposition, Lemma~\ref{tw:compr} immediately works for pathwidth.
Further, assume towards a contradiction that despite ETH, we can decide~$\Pi$ in time $2^{2^{o(k)}}\cdot\poly(\Card{\at(\Pi)})$. 
As in Theorem~\ref{thm:lb:fvs}, we take any 3-CNF formula~$F$, where~$\mathcal{G}_F$ is of width~$p$, and transform it into a normalized fully tight program~$\Pi'$.
From there, we compute a TD~$\mathcal{T}$ of~$\mathcal{G}_{{\Pi'}}$ of width~$p$, which can be done in time~$2^{\mathcal{O}(p^2)}\cdot\poly(\Card{\at(\Pi')})$ for pathwidth~$p$~\cite{KorhonenLokshtanov23}. From~$\mathcal{T}$, in polynomial time we construct a path decomposition~$\mathcal{P}$ of~$\mathcal{G}_{{\Pi'}}$, whose pathwidth is in~$\mathcal{O}(p)$~\cite{GroenlandEtAl23}.
Then, we apply our reduction and construct a normalized program~$\Pi$ of~$\mathcal{R}_{td}(\Pi', \mathcal{P})$, which runs in polynomial time. By Lemma~\ref{tw:compr}, $\mathcal{I}_{\Pi}$ admits a PD of width~$\mathcal{O}(k)$ with~$k\eqdef\log(p)$.
Solving~$\Pi$
in time $2^{2^{o(k)}}\cdot\poly(\Card{\at(\Pi')})$ contradicts ETH, similar to Theorem~\ref{thm:lb:fvs}.

The clique width lower bound immediately follows, since already a restricted version of cliquewidth, called linear cliquewidth, is linearly bounded by the pathwidth~\cite{FellowsEtAl09}. By the result above, we obtain the lower bound for linear cliquewidth, yielding the result for cliquewidth.
\end{proof}

\begin{restatelemma}[lem:pwbw]
\begin{lemma}
Let~$G{\,=\,}(V,E)$ be a graph, $\mathcal{T}{\,=\,}(T,\chi)$ be a PD of $G$ of width~$k$,
s.t.\ every vertex in~$V$ occurs in at most two  bags of~$\mathcal{T}$.
Then, the bandwidth of~$G$ is bounded by~$2k-1$.
\end{lemma}
\end{restatelemma}
\begin{proof}
We construct a vertex ordering~$f: V\rightarrow \{1,\ldots,\allowbreak\Card{V}\}$, by iterating over nodes in~$T$ from the leave towards the root.
First, for vertices in bag~$\chi(t')$ of leaf node~$t'$ in~$T$, we (arbitrarily) assign unique numbers $1,\ldots,\Card{\chi(t')}$ to the bag elements.
Then, for the subsequent bags~$\chi(t)$ with~$t'\in\children(t)$, we assign unique numbers~$n+1,\ldots,n+\Card{\chi(t)\setminus\chi(t')}$, where~$n$ is the largest number assigned for bag~$\chi(t')$ in the previous step, and so forth. 
Obviously, since every vertex in~$V$ appears in at most two neighboring bags, for every edge~$\{u,v\}$, value~$\Card{f(u)-f(v)}$ can be at most~$2k-1$, which bounds the bandwidth of~$G$.
\end{proof}

\begin{restatetheorem}[thm:bw]
\begin{theorem}[BW/CW LB]
Let~$\Pi$ be any normalized~disjunctive program. Under ETH, consistency of~$\Pi$ cannot be~decided in $2^{2^{o(k)}}\cdot\poly(\Card{\at(\Pi)})$ for bandwidth~$k$ of~$\mathcal{G}_\Pi$~($\mathcal{I}_\Pi$).
\end{theorem}
\end{restatetheorem}
\begin{proof}
First, we show the lower bound for~$\mathcal{G}_\Pi$.
We slightly strengthen the reduction~$\mathcal{R}_{td}$ such that we can define tree (path) decomposition, where every atom appears in at most~$2$ neighboring bags.
Let~$\Pi'$ be the given normalized fully tight program and $\mathcal{T}=(T,\chi,\varphi)$ be an annotated nice TD of~$\mathcal{G}_{\Pi'}$.
Then, the only problematic atom that appears in more than~$2$ neighboring bags in the proof construction of Lemma~\ref{tw:compr}, is~$sat_{\rootOf(T)}$. 
To fix this, we deploy copies~$sat'_t$ for every node~$t$ in~$T$
and for every~$t'\in\children(t)$, we add rules~$sat'_t \leftarrow sat'_{t'}$ and~$sat'_{t'} \leftarrow sat'_t$. 
These rules allow us to strengthen reduction~$\mathcal{R}_{td}$,
resulting in~$\mathcal{R}'_{td}$,
where we replace the occurrence of~$sat_{\rootOf(T)}$ in Rules~(\ref{redt:saturate})--(\ref{redt:bodyplus-false}
) by~$sat'_t$.
We use this adapted reduction, and let~$\Pi$ be a normalized program of $\mathcal{R}'_{td}(\Pi',\mathcal{T})$. Then, we can adapt the definition of~$\mathcal{T}'$ in Lemma~\ref{tw:compr}
to obtain a TD~$\mathcal{T}''$ of~$\mathcal{G}_{\Pi}$, such that every atom appears in at most~$2$ neighboring bags.
We define~$\mathcal{T}''\eqdef(T,\chi'')$, where for every~$t$ in~$T$, we let~$\chi''(t)\eqdef(\chi'(t)\setminus\{sat_{\rootOf(T)}\})\cup\{sat'_t, sat'_{t'}\}$, where~$\chi'$ is defined in Lemma~\ref{tw:compr}. Obviously, $\Card{\chi''(t)}$ is in~$\mathcal{O}(\log(k))$.

Since this constructions works for any TD, it also works for path decompositions. Then, by Lemma~\ref{lem:pwbw}, the bandwidth of~$\mathcal{G}_\Pi$ is in~$\mathcal{O}(\log(k))$. 
The lower bound then follows, using the same line of argumentation as in Theorem~\ref{thm:clw}.

The lower bound for the incidence graph~$\mathcal{I}_\Pi$  follows,
as one can further adapt~$\mathcal{T}''$, such that every bag contains at most one rule~$r\in\Pi$,
requiring node duplication in general.
\end{proof}

In order to obtain a tight bound for cutwidth, we require a constant degree.
We bound the degree as follows.

\begin{lemma}[Degree bound]\label{lem:bound}
Let~$\Pi$ be any normalized program, and~$\mathcal{T}$ be a nice TD/PD of~$\mathcal{G}_\Pi$ of width~$k$.
Then, there is a normalized program~$\Pi'$ such that (i) for every answer set~$M$ of~$\Pi$ there is a unique answer set~$M'$ of~$\Pi'$ and vice versa, (ii)
every vertex in~$\mathcal{G}_{\Pi'}$ ($\mathcal{I}_{\Pi'}$) has constant degree,
and (iii) there is a TD/PD~$\mathcal{T}'=(T',\chi')$ of~$\mathcal{G}_{\Pi'}$ ($\mathcal{I}_{\Pi'}$) of width~$2k$,
s.t.\ every atom in~$\at(\Pi')$ occurs in up to~$2$ bags.
%
\end{lemma}
\begin{proof}
First, we take~$\mathcal{T}$ and adapt it in polynomial time to obtain an annotated nice TD~$\mathcal{T}''=(T,\chi,\varphi)$ of~$\mathcal{G}_{\Pi}$.
Then, we construct a new program~$\Pi'$ as follows.
For every node~$t$ in~$T$ with~$t=\varphi(r^*)$ for~$r^*\in\Pi$,
we construct a copy~$r$ of the rule~$r^*$, where every atom occurrence~$a\in\at(r^*)$
is replaced by a fresh atom~$a_t$.
Further, for every node~$t$ in~$T$ with~$t'\in\children(t)$, we add the rules~$a_t \leftarrow a_{t'}$ and $a_{t'} \leftarrow a_t$ for encoding equivalence between copies.
Claim (i) holds: For any answer set~$M'$ of~$\Pi'$ there is a unique answer set~$M$ of~$\Pi$, constructed by removing $\cdot_t$ suffixes. Vice versa, one can uniquely construct~$M'$ given any answer set~$M$ of~$\Pi$ by replacing every atom~$a\in M$ by atoms of the form~$a_t$, for every node~$t$ in~$T$ with~$a\in\chi(t)$.
Further, (ii) is correct: Every atom in~$\mathcal{G}_{\Pi'}$ has at most degree $4$, up to $2$ neighbors are due to $\Pi$ being normalized (i.e., due to the constructed copy rules), and up to $2$ further neighbors are due to~$\mathcal{T}$ being nice and the constructed equivalence rules above.
Regarding (iii), we construct TD~$\mathcal{T}'\eqdef (T,\chi')$ such that for every node~$t$ of~$T$, we let~$\chi'(t)\eqdef \{a_t, a_t' \mid a\in \chi(t), t'\in\children(t)\}$. Since $\mathcal{T}'$ is a valid TD and $\Card{\chi'(t)} \leq 2\Card{\chi(t)}$, the claim~holds.

The result above can be adapted for the incidence graph, since (i) is unchanged, for (ii) vertex degrees are even bounded by~$3$ and for (iii) we can easily add to~$\chi'(t)$ the two equivalence rules for~$t$ and copy rule~$r\in\Pi'$ whenever~$t=\varphi(r^*)$ for the original rule~$r^*\in\Pi'$ of~$r$.

Further, since the result works for any TD, it immediately works for path decompositions as well.
\end{proof}

\begin{restatetheorem}[thm:cwlb]
\begin{theorem}[CW LB]
Let~$\Pi$ be any normalized disjunctive program. Under ETH the consistency of~$\Pi$ cannot be decided in time $2^{2^{o(k)}}\cdot\poly(\Card{\at(\Pi)})$ for cutwidth~$k$ of~$\mathcal{G}_\Pi$ or $\mathcal{I}_\Pi$.
\end{theorem}
\end{restatetheorem}
\begin{proof}
Let~$\Pi'$ be any normalized fully tight program and~$\mathcal{T}'$ be an annotated PD of~$\mathcal{G}_{\Pi'}$.
We construct a normalized fully tight program~$\Pi''$ from program~$\mathcal{R}_{td}(\Pi',\mathcal{T}')$. Then, by Lemma~\ref{lem:bound}, there is a program~$\Pi$ that bijectively preserves the answer sets of~$\Pi''$, where every atom has \emph{constant degree} in~$\mathcal{G}_\Pi$ ($\mathcal{I}_\Pi$). Further, by Lemma~\ref{lem:bound}, we also obtain a PD~$\mathcal{T}$ of~$\mathcal{G}_\Pi$ ($\mathcal{I}_\Pi$) such that every atom appears in at most two neighboring bags of~$\mathcal{T}$.
Since the cutwidth of a graph is bounded by its degree \emph{times} its bandwidth~\cite{KorachSolel93},
under ETH, $\Pi$ can not be decided in time $2^{2^{o(k)}}\cdot\poly(\Card{\at(\Pi)})$. Indeed, this yielded an algorithm, contradicting Theorem~\ref{thm:bw}.
%
\end{proof}

\end{document}